\documentclass{article}

\usepackage{microtype}
\usepackage{graphicx}
\usepackage{booktabs} 
\usepackage{subcaption}
\usepackage{verbatim}
\usepackage{enumitem}

\usepackage{hyperref}

\usepackage{amsmath,amsfonts,bm}
\usepackage{dsfont}

\def\eqref#1{equation~\ref{#1}}

\def\1{\bm{1}}

\def\vf{{\bm{f}}}

\def\vw{{\bm{w}}}
\def\vx{{\bm{x}}}
\def\vy{{\bm{y}}}

\DeclareMathAlphabet{\mathsfit}{\encodingdefault}{\sfdefault}{m}{sl}
\SetMathAlphabet{\mathsfit}{bold}{\encodingdefault}{\sfdefault}{bx}{n}

\newcommand{\mb}{\mathbf}
\newtheorem{definition}{Definition}
\graphicspath{ {figures/} }

\usepackage[accepted]{icml2021}

\icmltitlerunning{Are Bayesian neural networks intrinsically good at out-of-distribution detection?}

\usepackage{subfiles} 

\begin{document}

\twocolumn[
\icmltitle{Are Bayesian neural networks intrinsically good at\\out-of-distribution detection?}



\icmlsetsymbol{equal}{*}

\begin{icmlauthorlist}
\icmlauthor{Christian Henning}{equal,ini}
\icmlauthor{Francesco D'Angelo}{equal,ini}
\icmlauthor{Benjamin F. Grewe}{ini}
\end{icmlauthorlist}

\icmlaffiliation{ini}{Institute of Neuroinformatics, University of Zürich and ETH Zürich, Zürich, Switzerland}

\icmlcorrespondingauthor{Christian Henning}{\texttt{henningc@ethz.ch}}
\icmlcorrespondingauthor{Francesco D'Angelo}{fdangelo@student.ethz.ch}

\icmlkeywords{Machine Learning, ICML}

\vskip 0.3in
]



\printAffiliationsAndNotice{\icmlEqualContribution} 

\begin{abstract}
 The need to avoid confident predictions on unfamiliar data has sparked interest in \textit{out-of-distribution} (OOD) detection. It is widely assumed that Bayesian neural networks (BNN) are well suited for this task, as the endowed epistemic uncertainty should lead to disagreement in predictions on outliers. In this paper, we question this assumption and provide empirical evidence that proper Bayesian inference with common neural network architectures does not necessarily lead to good OOD detection. To circumvent the use of approximate inference, we start by studying the infinite-width case, where Bayesian inference can be exact considering the corresponding Gaussian process. Strikingly, the kernels induced under common architectural choices lead to uncertainties that do not reflect the underlying data generating process and are therefore unsuited for OOD detection. Finally, we study finite-width networks using HMC, and observe OOD behavior that is consistent with the infinite-width case. Overall, our study discloses fundamental problems when naively using BNNs for OOD detection and opens interesting avenues for future research.
\end{abstract}
\vspace{-1em}

\section{Introduction}
\label{sec:intro}
\begin{figure}[t]
\centering
\includegraphics[width=\linewidth, trim={0cm 0cm 0cm 0cm},clip]{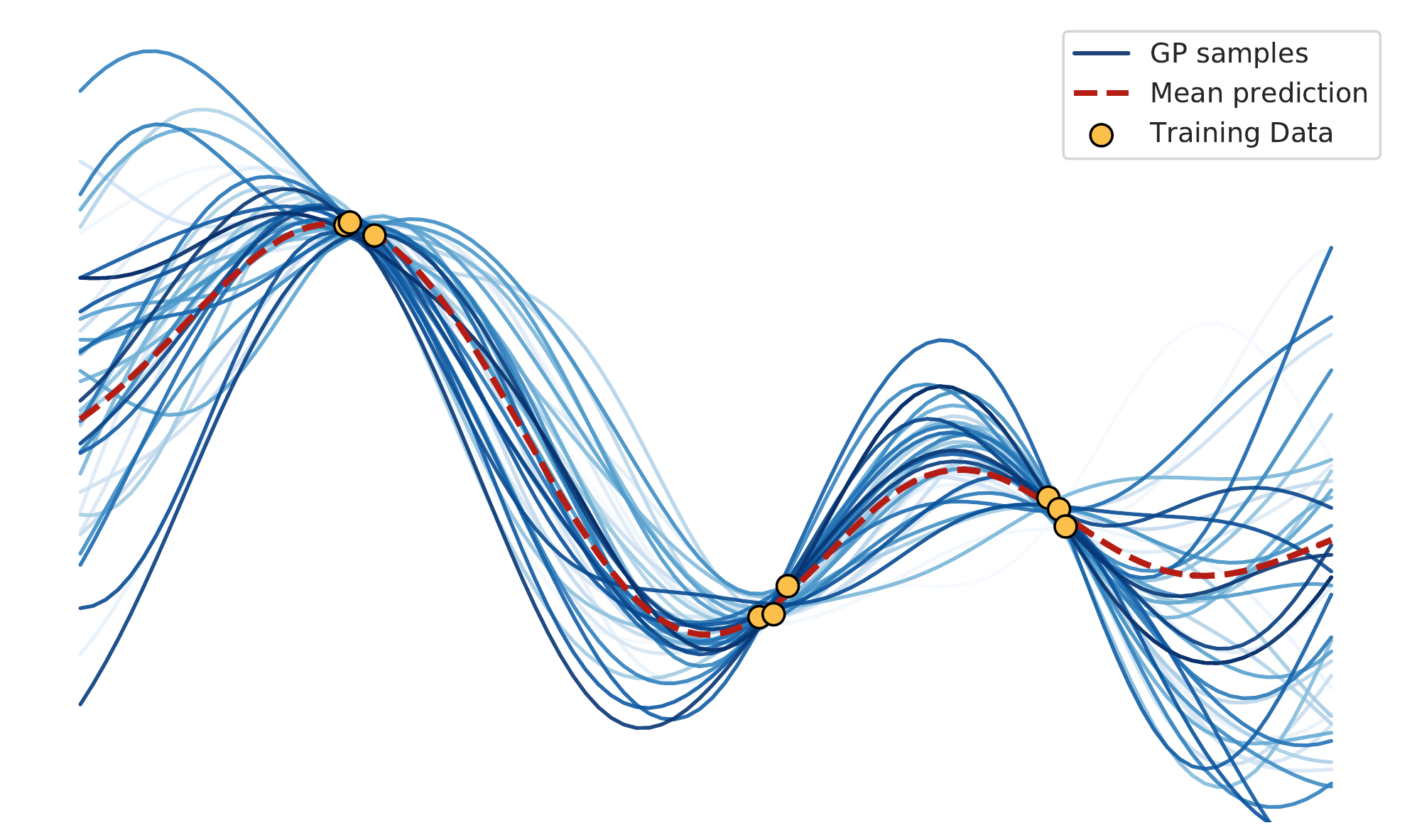}
\caption{\textbf{GP regression with an RBF kernel.} The prior variance over function values is only squeezed around training points, which leaves epistemic uncertainty high in OOD regions.}
\vspace{-1em}
\label{fig:gp_rbf}
\end{figure}

In supervised learning, the dataset is assumed to be generated according to some unknown process $\mathcal{D} \stackrel{i.i.d.}{\sim} p(\vx) p(\vy \mid \vx)$. The goal here is to infer from $\mathcal{D}$ alone the distribution $p(\vy \mid \vx)$ in order to make predictions on unseen inputs $\vx^*$. In the case of deep learning, this problem is approached by choosing a neural network $f(\cdot; \vw)$ parametrized by $\vw$, and predictions are made via the conditional $p\big(\vy \mid f(\cdot; \vw)\big)$ (also called the \textit{likelihood}). Assuming that the induced class of models contains some $\hat{\vw}$ such that $p\big(\vy \mid f(\cdot; \hat{\vw})\big) = p(\vy \mid \vx)$ almost everywhere on the support of $p(\vx)$, Bayesian statistics can be used to infer plausible models $p(\vw \mid \mathcal{D})$ under the observed data given some prior knowledge $p(\vw)$ (see \citet{mackay2003itila} for an introduction).

Bayesian inference has a multitude of benefits as it is not susceptible to overfitting and new evidence can be incorporated without requiring access to past data. Furthermore, the explicit handling of parameter uncertainty is even more important in the overparametrized regime where the increased number of degrees of freedom further extends the hypothesis class, providing an intuitive need of a Bayesian treatment, but also deceptively motivates the naive use of BNNs as OOD detectors.
The issue with this argument is partly rooted in the fact, that there is no universal mathematical definition that characterizes an unseen input $\vx^*$ as out-of-distribution (see SM \ref{sm:sec:ood} for a detailed discussion). In this work, we boldly assume that the observed data $\mathcal{D}$ allows us to generalize well and mimic $p(\vy \mid \vx)$ across the high-probability regions of $p(\vx)$. In this case, meaningful predictions can be made unless an input $\vx^*$ is unlikely under $p(\vx)$ or even from outside the support of $p(\vx)$. These are the type of inputs we consider OOD, and providing guarantees for their  detection is of utmost importance for safety-critical applications of AI and reliable predictions.

While it appears natural to tackle this problem from a generative perspective by explicitly modelling $p(\vx)$, this paper is solely concerned with the question of how justified it is to deploy uncertainty-based OOD detection via a Bayesian neural network.

The uncertainty captured by a BNN can be coarsely categorized into \textit{aleatoric} and \textit{epistemic} uncertainty. Aleatoric uncertainty is irreducible and intrinsic to the data $p(\vy \mid \vx)$. For instance, a blurry image might not contain enough information to identify unique object classes. On the other hand, a BNN also models epistemic uncertainty by maintaining a distribution over parameters $p(\vw \mid \mathcal{D})$. This distribution reflects the uncertainty about which hypothesis explains the data and can be reduced by observing more data. Arguably, aleatoric uncertainty is of little interest for detecting OOD inputs, while epistemic uncertainty might be useful if the following assumption holds: the hypotheses captured by $p(\vw \mid \mathcal{D})$ must agree on their predictions for in-distribution samples but disagree for OOD samples (Fig. \ref{fig:gp_rbf}). Certain Bayesian methods satisfy this assumption, e.g. Gaussian process regression with an RBF kernel (cf. Sec. \ref{sec:discussion}). However, the uncertainty induced by Bayesian inference does not in general give rise to OOD capabilities. This can easily be verified by considering the following thought experiment: Assume a model class with only two hypotheses $g(x)$ and $h(x)$ such that $g(x) \neq h(x) \forall x$ and data being generated according to $g(x)$. Once data is observed, we can commit to the ground-truth hypothesis $g(x)$ and thus lose epistemic uncertainty in- and out-of-distribution. Finite-width neural networks on the other hand, form a powerful class of models, and are often put into context with universal function approximators \cite{hornik:1991:uat}. But is this fact enough to attribute them with good OOD capabilities? The literature often seems to imply that a BNN is intrinsically good at OOD detection. For instance, the use of OOD benchmarks when introducing new methods for approximate inference creates the false impression that the true posterior is a good OOD detector \cite{louizos:multiplicative:nf:2017, pawlowski2017implicit:weight:uncertainty, krueger_bayesian_2017, henning2018approximating, nips:2019:maddox, ciosek:2020:conservative:uncertainty:estimation}. Our work is meant to kickstart a discussion among researchers about the properties of OOD uncertainty when performing exact Bayesian inference in neural networks. To achieve this, we contribute as follows:
\vspace{-1em}

\begin{itemize}[leftmargin=*]
    \setlength\itemsep{.1em}
    \item We empirically show that exact inference in infinite-width networks does not necessarily lead to desirable OOD behavior, and that these observations are consistent with results obtained via approximate inference in their finite-width counterparts.
    \item We study the OOD behavior of infinite-width networks on regression tasks by analysing the properties of the induced kernels. Moreover, we discuss desirable kernel features for OOD detection and empirically show that these are not present in common architectural choices. 
    \item We emphasize that the choice of weight-space prior has a strong effect on OOD performance.
    \item Finally, we discuss uncertainty quantification, highlighting that the commonly used predictive entropy is confounded by aleatoric uncertainty.
\end{itemize}
\vspace{-1em}

\section{Background}
\label{sec:background}

In this section, we briefly introduce the concepts on which we base our  argumentation in Sec. \ref{sec:discussion}. We start by introducing BNNs, which unfortunately rely on approximate inference. However, in the non-parametric limit and under a certain choice of prior, a BNN converges to a Gaussian process, an alternative Bayesian inference framework where exact inference is possible. The connection between BNNs and Gaussian processes will later allow us to make interesting conjectures about OOD behavior. 

\textbf{Bayesian neural networks.}
In supervised deep learning, we typically consider  a likelihood function $p(\vy \mid  f(\vx; \mb{w}))$ parameterized by a neural network $f(\vx; \vw)$ and training data $\mathcal{D} = \{(\vx_i,\vy_i)\}_{i=1}^n$. 
In BNNs, we are interested in the  posterior distribution of all likely networks given by:
\begin{equation}
    p(\vw \mid  \mathcal{D}) \propto \prod_{i=1}^n p\left(\vy_i \mid  f(\vx_i ; \vw ) \right) \, p(\vw)
    \label{eqn:BNN_post}
\end{equation}
where $p(\vw)$ is the prior distribution over weights.
Crucially, when making a prediction on a test point $\vx^*$, in the Bayesian approach we do not only use a single parameter $\hat{\vw}$ but we marginalize over the whole posterior, thus taking all possible explanations of the data into account:
\begin{equation}
    p(\vy^* \mid  \vx^*, \mathcal{D}) = \int p(\vy^* \mid  f(\vx^*; \vw)) \, p(\vw \mid  \mathcal{D}) \, \mathrm{d}\vw
    \label{eqn:BMA}
\end{equation}

\textbf{Gaussian process.}
Gaussian processes are established Bayesian machine learning models that, despite their strong scalability limitations, can offer a powerful inference framework. Formally \cite{rasmussen2004gp}: 
\vspace{-.5em}
\begin{definition}
A Gaussian process (GP) is a collection of random variables, any finite number of which have a joint Gaussian distribution. 
\end{definition}
\vspace{-.5em}
Compared to parametric models, GPs have the advantage of performing inference directly in function space. A GP is defined by its mean $m(\vx) = \mathbb{E}\left[\vf(\vx)\right]$ and covariance $C\left( \vf(\vx), \vf(\vx')\right)=\mathbb{E}\left[\left(\vf(\vx) - m(\vx)\right) \left(\vf(\vx')- m(\vx')\right)\right]$.
The latter can be specified using a kernel $k: \mathbb{R}^d\times  \mathbb{R}^d \to \mathbb{R}$ and implies a prior distribution over functions: 
\begin{equation}
    p(\vf|X) = \mathcal{N}\left(\mathbf{0}, K(X,X) \right) \, ,
    \label{eq:gp:prior}
\end{equation}
where $K(X,X)_{ij} = k(\vx_i,\vx_j)$ is the kernel Gram matrix on the training inputs $X$, and $m(\vx)$ has been chosen to be $0$.  
When observing the training data, the prior is reshaped to place more mass in the regions of functions that are more likely to have generated them, and this knowledge is then used to make predictions on unseen inputs $X_\star$. In probabilistic terms, this operation corresponds to conditioning the joint Gaussian prior: $ p(\vf_*|X_*,X, \vf)$. To model the data more realistically, we assume to not have direct access to the function values but to noisy observations: $\vy = \vf(
\vx) + \epsilon$ with $\epsilon \sim \mathcal{N}(0,\sigma^2 \mathbb{I}_d)$. This assumption is formally equivalent to a Gaussian likelihood $p(\vy|\mb{f}) = \mathcal{N}(\vy|\vf,\sigma^2)$. The conditional distribution on the noisy observation can then be written as \cite{rasmussen2004gp}:
\begin{equation}
    \begin{split}
         p(\vf_* \mid X_*,X, \vy) &= \int d \vf p(\vf_* \mid X_*,X, \vf) p(\vf \mid X, \vy) \\&= \mathcal{N}\left(\bar{\vf}_*, C(\vf_*) \right) \\ 
 \bar{\vf}_* &= K(X_*,X)[K(X,X) + \sigma^2 \mathbb{I}]^{-1} \vy \\ 
  C(\vf_*) &= K(X_*,X_*) - K(X_*,X)[K(X,X) \\ &\hspace{22mm}+ \sigma^2 \mathbb{I}]^{-1}K(X,X_*)
    \end{split}
    \label{eqn:gp_regr}
\end{equation}
where $p(\vf \mid X, \vy) = \frac{p(\vy \mid \vf ) p(\vf \mid X))}{p(\vy \mid X)}$.

\textbf{The relation of infinite-width BNNs and GPs.}
The connection between neural networks and GPs has recently gained significant attention. \citet{neal1996bayesian} showed that a 1-hidden layer BNN converges to a GP in the infinite-width limit. More recently, the work by \citet{lee2018deep} extended this result to deeper networks, called \textit{neural network Gaussian processes} (NNGP). Crucially, the kernel function of the related GP strictly depends on the used activation function. To better understand this connection we consider a fully connected network with $L$ hidden layers $l=1,\dots,L$ with width $H_l$. For each input we use $\mb{x}^l(\vx)$ to represent the post-activation with $\mb{x}^0 = \vx$ and $\vf^l$ the pre-activation so that $f_i^l(\vx) = b^l_i + \sum_{j=1}^{H_l} w_{ij}^l x^l_j$ with $x^l_j = h(f_j^{l-1})$ and $h(\cdot)$ a point-wise activation function. Furthermore, the weights and biases are distributed according to $b_j^l \sim \mathcal{N}(0,\sigma_b^2)$ and $w_{ij}^l \sim \mathcal{N}(0,\frac{\sigma_w^2}{H_l})$. Given the independence of the weights and biases also the post-activations are independent so that the central limit theorem can be applied, and for $H_l \to \infty $ we obtain $f^l(\vx) \sim \mathcal{GP} (0,C^l)$. The covariance $C^l$ is specified by the kernel induced by the network: 
\begin{equation}
    \begin{split}
        C^l(\vx,\vx') &= \overbrace{\mathbb{E} \left[f_i^l(\vx) f_i^l(\vx') \right]}^{k^l(\vx,\vx')} - \overbrace{\mathbb{E} [f^l_i(\vx)]  \mathbb{E} [f^l_i(\vx)]}^{=0} \\ 
        &= \sigma_b^2 + \sigma^2_w\mathbb{E}_{f_i^{l-1}} \left[h\left(f_i^{l-1}(\vx)\right) h\left(f_i^{l-1}(\vx') \right) \right] \, .
    \label{eqn:MC_NNGP}
    \end{split}
\end{equation}
For the first layer we have:
\begin{equation}
    k^0(\vx,\vx') = \mathbb{E} \left[f_i^0(\vx) f_i^0(\vx') \right] = \sigma_b^2 + \sigma_w^2 \left( \frac{\vx^T \vx'}{d} \right) \, .
    \label{eqn:k:recursion:start}
\end{equation}
When using a ReLU activation function, the kernel has a closed form solution \cite{relu:kernel:cho}: 
\begin{equation}
\begin{split}
     k^l_{ReLU}(\vx,\vx') &= \sigma_b^2 + \frac{\sigma_w^2}{2 \pi} \sqrt{k^{l-1}(\vx,\vx)k^{l-1}(\vx',\vx')} \\ &\left( \sin \theta_{\vx,\vx'}^{l-1} + (\pi - \theta^{l-1}_{\vx,\vx'}) \cos \theta^{l-1}_{\vx,\vx'}\right)  \\
     \theta_{\vx,\vx'}^l &= \cos^{-1} \left( \frac{k^l(\vx,\vx')}{\sqrt{k^l(\vx,\vx)k^l(\vx',\vx')}}\right)
    \label{eqn:k_relu}
\end{split}
\end{equation}
Whereas activations like sigmoid or hyperbolic tangent, require a Monte Carlo estimate of Eq.~\ref{eqn:MC_NNGP} (cf. Fig. \ref{sm:fig:mc:error}). 
\vspace{-1em}

\section{Discussion and empirical observations}
\label{sec:discussion}

In the previous section, we recalled the connection between BNNs and GPs, namely that Bayesian inference in an infinite-width neural network can be studied in the GP framework (assuming a proper choice of prior). This connection allows studying how the kernels induced by architectural choices shape the prior in function space, and how these choices ultimately determine OOD behavior. In this section, we analyze this OOD behavior for traditional as well as NNGP-induced kernels. To minimize the impact of the approximations on our results, we exclusively focus on  conjugate settings, i.e., regression.

\begin{figure*}
     \centering
     \begin{subfigure}[b]{0.24\textwidth}
         \centering
         \includegraphics[width=\textwidth]{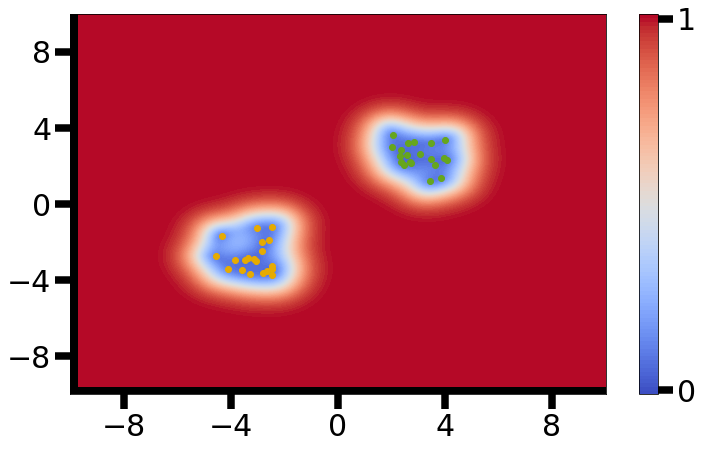}
         \caption{$\sigma(\vf_*)$ -- RBF kernel}
         \label{fig:gp:rbf:std}
     \end{subfigure}
     \hfill
     \begin{subfigure}[b]{0.24\textwidth}
         \centering
         \includegraphics[width=\textwidth]{gauss_mix/nngp/post_std_ana_relu_2l.png}
         \caption{$\sigma(\vf_*)$ -- 2-layer ReLU ($\infty$)}
         \label{fig:gp:ana:relu:nngp:2l}
     \end{subfigure}
     \hfill
     \begin{subfigure}[b]{0.247\textwidth}
         \centering
         \includegraphics[width=\textwidth]{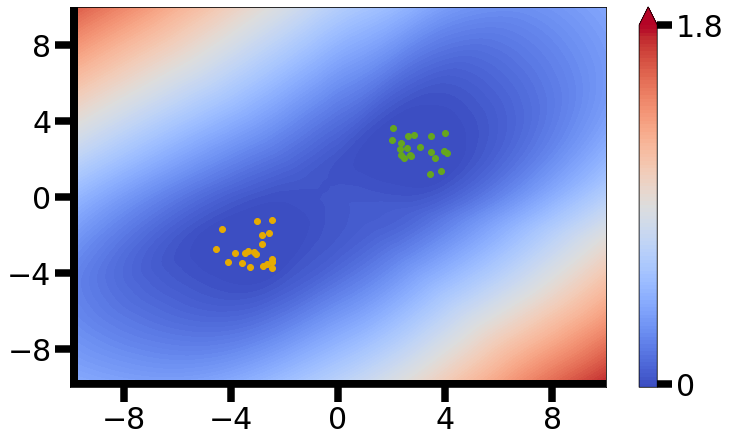}
         \caption{$\sigma(\vf_*)$ -- 2-layer ReLU (5)}
         \label{fig:hmc:std:relu:2l:width:10}
     \end{subfigure}
     \begin{subfigure}[b]{0.247\textwidth}
         \centering
         \includegraphics[width=\textwidth]{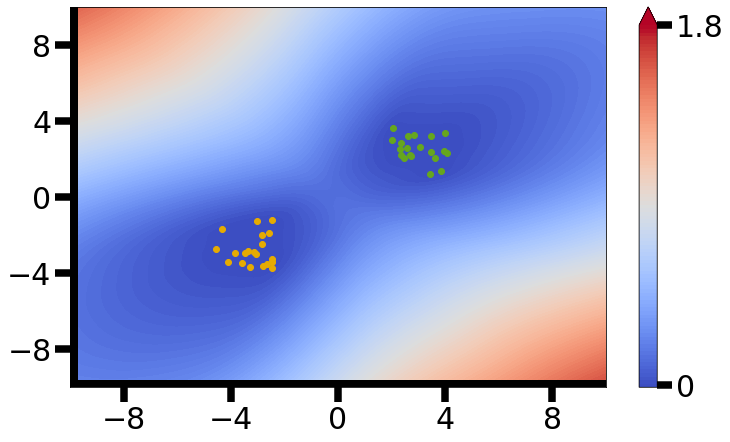}
         \caption{$\sigma(\vf_*)$ -- 2-layer ReLU (100)}
         \label{fig:hmc:std:relu:2l:width:100}
     \end{subfigure}
        \caption{\textbf{Standard deviation of the predictive posterior.} We perform Bayesian inference on a mixture of two Gaussians dataset considering different priors in function space. The problem is treated as regression task to allow exact inference in combination with GPs.}
        \label{fig:main:ood:plots}
\end{figure*}

\textbf{GP regression with an RBF kernel.}
We first examine the predictive posterior of a GP with squared exponential (RBF) kernel $k(\vx_p, \vx_q) = \exp \big(- \frac{1}{2l^2} \lVert\vx_p - \vx_q \rVert^2_2 \big)$, where $l$ is a hyperparameter. Fig. \ref{fig:gp:rbf:std} shows the standard deviation $\sigma(\vf_*) \equiv \sqrt{\text{diag} \left( C(\vf_*) \right)}$ of the predictive posterior. We can notice that $\sigma(\vf_*)$ nicely captures the data manifold and is thus well suited for OOD detection. This behavior can be understood by considering Eq. \ref{eqn:gp_regr} while noting that $k(\vx_p, \vx_q) = \text{const}$ if $\vx_p = \vx_q$. In this case, the variance of posterior function values can be written as $\sigma^2(\vf_*) = k(x_*,x_*) - \sum_{i=1}^n \beta_i(x_*) k(x_*,x_i)$ (cf. SM \ref{sm:sec:rbf:kde}), where $\beta_i$ are dataset- and input-dependent. The second term is reminiscent of using kernel density estimation (KDE) to approximate $p(\vx)$, applying a Gaussian kernel, while the first term is the (constant) prior variance. Hence, the link between Bayesian inference and OOD detection can be made explicit, as the posterior variance is inversely related to the input distribution. In this view, one starts with high (prior) uncertainty everywhere, which is only reduced where data is observed. By contrast, learning a normalized generative model (e.g., using normalizing flows \cite{papamakarios2019nf:review}) often requires to start from an arbitrary probability distribution. Importantly, the desirable OOD behavior observed for RBFs is not generalizable to all kernels (cf. Fig. \ref{sm:fig:traditional:kernels}).

\textbf{Uncertainty quantification for OOD detection.} Uncertainty can be quantified in multiple ways, but is often measured as the entropy of the predictive posterior. The predictive posterior, however, captures both aleatoric and epistemic uncertainty, which does not allow a distinction between OOD and ambiguous inputs \cite{gal:2021:deterministic:etpistemic:aleatoric}. While our choice of likelihood does not permit the modelling of input-dependent uncertainty (Gaussian with fixed variance), a softmax classifier can capture aleatoric uncertainty arbitrarily well. For this reason, uncertainty should be quantified in a way that allows OOD detection to be based on epistemic uncertainty only.

The function space view of GPs naturally provides such measure of epistemic uncertainty by considering the posterior variance of function values $\sigma^2(\vf_*)$ (illustrated in Fig. \ref{fig:gp_rbf}). Such measure can be naturally translated to BNNs by looking at the disagreement between network outputs when sampling from $p(\vw \mid  \mathcal{D})$. Note that in classification tasks all models drawn from $p(\vw \mid  \mathcal{D})$ might lead to a high-entropy softmax without disagreement, and therefore only an uncertainty measure based on model disagreement prevents one from misjudging ambiguous points as OOD. As OOD detection is the focus of this paper, we always quantify uncertainty in terms of model disagreement.



\textbf{The OOD behavior induced by NNGP kernels.}
Fig. \ref{fig:gp:ana:relu:nngp:2l} illustrates $\sigma^2(\vf_*)$ for an infinite-width 2-layer ReLU network (see Fig. \ref{sm:fig:nngp:posteriors} for other architectural choices). It is already visually apparent, that in this case the kernel is less suited for OOD detection compared to an RBF kernel. Moreover, we cannot justify why OOD detection based on $\sigma^2(\vf_*)$ would be principled for this kernel as the KDE analogy does not hold. This is due to two reasons: (1) the prior uncertainty $k(x_*,x_*)$ is not constant (Fig. \ref{sm:fig:prior:kernels}), and (2) we empirically do not observe that $k(\vx_p, \vx_q)$ can be related to a distance measure (Fig. \ref{sm:fig:nngp:distance:awareness}). We therefore argue that more theoretical work is necessary to justify the use of NNGP kernels in the context of OOD detection.

On this note, maintaining parameter uncertainty and being able to detect OOD samples are often considered crucial requirements of systems deployed in safety-critical applications. Given that Bayesian inference is not intrinsically linked to OOD detection, care should be taken to precisely communicate safety-relevant capabilities of BNNs to practitioners.

\textbf{Infinite-width uncertainty is consistent with the finite-width uncertainty.} We next consider finite-width BNNs, where exact Bayesian inference is intractable. To mitigate the effects of approximate inference we resort to Hamiltonian Monte Carlo \citep[HMC, ][]{duane1987hmc, neal:2011:hmc}. Fig. \ref{fig:hmc:std:relu:2l:width:10} and \ref{fig:hmc:std:relu:2l:width:100} show an estimate of $\sigma^2(\vf_*)$ for finite-width networks.
Already for moderate layer widths, the uncertainty modelled resembles the one of the corresponding NNGP. Given this close correspondence, we conjecture that the tools available for the infinite-width case are useful for designing architectural guidelines that enhance OOD detection. We explore this idea using RBF networks in SM \ref{sm:sec:rbf:nets}.

\textbf{The importance of the weight-space prior.} The correspondence between NNGPs and BNNs only applies for zero-mean priors where the variance is corrected by the layer width. However, the combination of architecture and $p(\vw)$ induces a prior over functions \cite{wilson2020bayesian:generalization}, which will ultimately shape posterior's OOD uncertainty. We highlight the importance of $p(\vw)$ in Fig.~\ref{sm:fig:prior:importance}. Note, that studying GP kernels with respect to their OOD performance can also help designing meaningful weight-space priors for a given architecture, as $p(\vw)$ can be learned to induce a prior in function space that mimics a GP prior \citep[e.g., ][]{flam:2017:gp:prior:bnn}.

\textbf{Conclusion.} We discussed the relation of BNNs with OOD detection from a GP perspective. Due to computational restrictions, our empirical results are limited to low-dimensional problems. However, the fact that empirically BNNs are often only marginally superior in OOD detection compared to models that do not maintain epistemic uncertainty \cite{snoek2019can:you:trust:your:models:uncertainty,posterior:replay:2021:henning:cervera}, indicates fundamental limitations that are not solely explained by the use of approximate inference.

Overall, this work raises doubts about the capabilities of BNNs for OOD detection and challenges the community to provide evidence that justifies the wide-spread use for that purpose.

\section*{Acknowledgements}

This work was supported by the Swiss National Science Foundation (B.F.G. CRSII5-173721 and 315230\_189251), ETH project funding (B.F.G. ETH-20 19-01) and funding from the Swiss Data Science Center (B.F.G, C17-18). In particular, we would like to thank Maria R. Cervera for countless fruitful discussions and her assistance in writing the manuscript. Furthermore, we would like to thank Benjamin Ehret, João Sacramento, Simone Carlo Surace and Jean-Pascal Pfister for discussions and feedback.

\bibliography{main}

\begin{thebibliography}{28}
\providecommand{\natexlab}[1]{#1}
\providecommand{\url}[1]{\texttt{#1}}
\expandafter\ifx\csname urlstyle\endcsname\relax
  \providecommand{\doi}[1]{doi: #1}\else
  \providecommand{\doi}{doi: \begingroup \urlstyle{rm}\Url}\fi

\bibitem[Cho \& Saul(2009)Cho and Saul]{relu:kernel:cho}
Cho, Y. and Saul, L.
\newblock Kernel methods for deep learning.
\newblock In Bengio, Y., Schuurmans, D., Lafferty, J., Williams, C., and
  Culotta, A. (eds.), \emph{Advances in Neural Information Processing Systems},
  volume~22. Curran Associates, Inc., 2009.

\bibitem[Ciosek et~al.(2020)Ciosek, Fortuin, Tomioka, Hofmann, and
  Turner]{ciosek:2020:conservative:uncertainty:estimation}
Ciosek, K., Fortuin, V., Tomioka, R., Hofmann, K., and Turner, R.
\newblock Conservative uncertainty estimation by fitting prior networks.
\newblock In \emph{International Conference on Learning Representations}, 2020.

\bibitem[Duane et~al.(1987)Duane, Kennedy, Pendleton, and Roweth]{duane1987hmc}
Duane, S., Kennedy, A.~D., Pendleton, B.~J., and Roweth, D.
\newblock Hybrid monte carlo.
\newblock \emph{Physics letters B}, 195\penalty0 (2):\penalty0 216--222, 1987.

\bibitem[Flam-Shepherd et~al.(2017)Flam-Shepherd, Requeima, and
  Duvenaud]{flam:2017:gp:prior:bnn}
Flam-Shepherd, D., Requeima, J., and Duvenaud, D.
\newblock Mapping gaussian process priors to bayesian neural networks.
\newblock In \emph{NIPS Bayesian deep learning workshop}, 2017.

\bibitem[Gretton et~al.(2012)Gretton, Borgwardt, Rasch, Sch{{\"o}}lkopf, and
  Smola]{kernel:two:sample:test}
Gretton, A., Borgwardt, K.~M., Rasch, M.~J., Sch{{\"o}}lkopf, B., and Smola, A.
\newblock A kernel two-sample test.
\newblock \emph{Journal of Machine Learning Research}, 13\penalty0
  (25):\penalty0 723--773, 2012.

\bibitem[Henning et~al.(2018)Henning, von Oswald, Sacramento, Surace, Pfister,
  and Grewe]{henning2018approximating}
Henning, C., von Oswald, J., Sacramento, J., Surace, S.~C., Pfister, J.-P., and
  Grewe, B.~F.
\newblock Approximating the predictive distribution via adversarially-trained
  hypernetworks.
\newblock In \emph{NeurIPS Bayesian Deep Learning Workshop}, 2018.

\bibitem[Henning et~al.(2021)Henning, Cervera, D'Angelo, von Oswald, Traber,
  Ehret, Kobayashi, Sacramento, and
  Grewe]{posterior:replay:2021:henning:cervera}
Henning, C., Cervera, M.~R., D'Angelo, F., von Oswald, J., Traber, R., Ehret,
  B., Kobayashi, S., Sacramento, J., and Grewe, B.~F.
\newblock Posterior meta-replay for continual learning.
\newblock \emph{arXiv}, 2021.

\bibitem[Hornik(1991)]{hornik:1991:uat}
Hornik, K.
\newblock Approximation capabilities of multilayer feedforward networks.
\newblock \emph{Neural Networks}, 4\penalty0 (2):\penalty0 251--257, 1991.
\newblock ISSN 0893-6080.
\newblock \doi{https://doi.org/10.1016/0893-6080(91)90009-T}.

\bibitem[Krueger et~al.(2017)Krueger, Huang, Islam, Turner, Lacoste, and
  Courville]{krueger_bayesian_2017}
Krueger, D., Huang, C.-W., Islam, R., Turner, R., Lacoste, A., and Courville,
  A.
\newblock Bayesian {Hypernetworks}.
\newblock \emph{arXiv:1710.04759 [cs, stat]}, October 2017.
\newblock arXiv: 1710.04759.

\bibitem[Lee et~al.(2018)Lee, Sohl-dickstein, Pennington, Novak, Schoenholz,
  and Bahri]{lee2018deep}
Lee, J., Sohl-dickstein, J., Pennington, J., Novak, R., Schoenholz, S., and
  Bahri, Y.
\newblock Deep neural networks as gaussian processes.
\newblock In \emph{International Conference on Learning Representations}, 2018.

\bibitem[Liu et~al.(2016)Liu, Lee, and Jordan]{kernalized:stein:discrepancy}
Liu, Q., Lee, J., and Jordan, M.
\newblock A kernelized stein discrepancy for goodness-of-fit tests.
\newblock In Balcan, M.~F. and Weinberger, K.~Q. (eds.), \emph{Proceedings of
  The 33rd International Conference on Machine Learning}, volume~48 of
  \emph{Proceedings of Machine Learning Research}, pp.\  276--284, New York,
  New York, USA, 20--22 Jun 2016. PMLR.

\bibitem[Louizos \& Welling(2017)Louizos and
  Welling]{louizos:multiplicative:nf:2017}
Louizos, C. and Welling, M.
\newblock Multiplicative {Normalizing} {Flows} for {Variational} {Bayesian}
  {Neural} {Networks}.
\newblock In \emph{Proceedings of the 34th {International} {Conference} on
  {Machine} {Learning} - {Volume} 70}, {ICML}'17, pp.\  2218--2227. JMLR.org,
  2017.

\bibitem[MacKay(2003)]{mackay2003itila}
MacKay, D.~J.
\newblock \emph{Information theory, inference and learning algorithms}.
\newblock Cambridge university press, 2003.

\bibitem[MacKay(2002)]{mackay:information:theory}
MacKay, D. J.~C.
\newblock \emph{Information Theory, Inference and Learning Algorithms}.
\newblock Cambridge University Press, USA, 2002.
\newblock ISBN 0521642981.

\bibitem[Maddox et~al.(2019)Maddox, Izmailov, Garipov, Vetrov, and
  Wilson]{nips:2019:maddox}
Maddox, W.~J., Izmailov, P., Garipov, T., Vetrov, D.~P., and Wilson, A.~G.
\newblock A simple baseline for bayesian uncertainty in deep learning.
\newblock In Wallach, H., Larochelle, H., Beygelzimer, A., d\textquotesingle
  Alch\'{e}-Buc, F., Fox, E., and Garnett, R. (eds.), \emph{Advances in Neural
  Information Processing Systems}, volume~32. Curran Associates, Inc., 2019.

\bibitem[Mukhoti et~al.(2021)Mukhoti, Kirsch, van Amersfoort, Torr, and
  Gal]{gal:2021:deterministic:etpistemic:aleatoric}
Mukhoti, J., Kirsch, A., van Amersfoort, J., Torr, P.~H., and Gal, Y.
\newblock Deterministic neural networks with appropriate inductive biases
  capture epistemic and aleatoric uncertainty.
\newblock \emph{arXiv}, 2021.

\bibitem[Nalisnick et~al.(2019{\natexlab{a}})Nalisnick, Matsukawa, Teh, Gorur,
  and Lakshminarayanan]{nalisnick:2018ood:generative}
Nalisnick, E., Matsukawa, A., Teh, Y.~W., Gorur, D., and Lakshminarayanan, B.
\newblock Do deep generative models know what they don't know?
\newblock In \emph{International Conference on Learning Representations},
  2019{\natexlab{a}}.

\bibitem[Nalisnick et~al.(2019{\natexlab{b}})Nalisnick, Matsukawa, Teh, and
  Lakshminarayanan]{nalisnick2019typical}
Nalisnick, E., Matsukawa, A., Teh, Y.~W., and Lakshminarayanan, B.
\newblock Detecting out-of-distribution inputs to deep generative models using
  typicality.
\newblock \emph{arXiv preprint arXiv:1906.02994}, 2019{\natexlab{b}}.

\bibitem[Neal(1996)]{neal1996bayesian}
Neal, R.~M.
\newblock \emph{Bayesian Learning for Neural Networks}.
\newblock Springer-Verlag, Berlin, Heidelberg, 1996.
\newblock ISBN 0387947248.

\bibitem[Neal et~al.(2011)]{neal:2011:hmc}
Neal, R.~M. et~al.
\newblock Mcmc using hamiltonian dynamics.
\newblock \emph{Handbook of markov chain monte carlo}, 2\penalty0
  (11):\penalty0 2, 2011.

\bibitem[Papamakarios et~al.(2019)Papamakarios, Nalisnick, Rezende, Mohamed,
  and Lakshminarayanan]{papamakarios2019nf:review}
Papamakarios, G., Nalisnick, E., Rezende, D.~J., Mohamed, S., and
  Lakshminarayanan, B.
\newblock Normalizing flows for probabilistic modeling and inference.
\newblock \emph{arXiv}, 2019.

\bibitem[Pawlowski et~al.(2017)Pawlowski, Brock, Lee, Rajchl, and
  Glocker]{pawlowski2017implicit:weight:uncertainty}
Pawlowski, N., Brock, A., Lee, M.~C., Rajchl, M., and Glocker, B.
\newblock Implicit weight uncertainty in neural networks.
\newblock \emph{arXiv}, 2017.

\bibitem[Pimentel et~al.(2014)Pimentel, Clifton, Clifton, and
  Tarassenko]{pimentel2014review:novelty}
Pimentel, M.~A., Clifton, D.~A., Clifton, L., and Tarassenko, L.
\newblock A review of novelty detection.
\newblock \emph{Signal Processing}, 99:\penalty0 215--249, 2014.

\bibitem[Rasmussen(2004)]{rasmussen2004gp}
Rasmussen, C.~E.
\newblock \emph{Gaussian Processes in Machine Learning}, pp.\  63--71.
\newblock Springer Berlin Heidelberg, Berlin, Heidelberg, 2004.
\newblock ISBN 978-3-540-28650-9.
\newblock \doi{10.1007/978-3-540-28650-9_4}.

\bibitem[Snoek et~al.(2019)Snoek, Ovadia, Fertig, Lakshminarayanan, Nowozin,
  Sculley, Dillon, Ren, and
  Nado]{snoek2019can:you:trust:your:models:uncertainty}
Snoek, J., Ovadia, Y., Fertig, E., Lakshminarayanan, B., Nowozin, S., Sculley,
  D., Dillon, J., Ren, J., and Nado, Z.
\newblock Can you trust your model's uncertainty? evaluating predictive
  uncertainty under dataset shift.
\newblock In \emph{Advances in Neural Information Processing Systems}, pp.\
  13969--13980, 2019.

\bibitem[Williams(1997)]{williams1997rbf}
Williams, C.~K.
\newblock Computing with infinite networks.
\newblock \emph{Advances in neural information processing systems}, pp.\
  295--301, 1997.

\bibitem[Wilson \& Izmailov(2020)Wilson and
  Izmailov]{wilson2020bayesian:generalization}
Wilson, A.~G. and Izmailov, P.
\newblock Bayesian deep learning and a probabilistic perspective of
  generalization.
\newblock In \emph{Advances in Neural Information Processing Systems}, 2020.

\bibitem[Zimek \& Filzmoser(2018)Zimek and
  Filzmoser]{zimek2018outlier:detection}
Zimek, A. and Filzmoser, P.
\newblock There and back again: Outlier detection between statistical reasoning
  and data mining algorithms.
\newblock \emph{Wiley Interdisciplinary Reviews: Data Mining and Knowledge
  Discovery}, 8\penalty0 (6):\penalty0 e1280, 2018.

\end{thebibliography}
\bibliographystyle{icml2021}


\newpage
\normalsize
\setcounter{page}{1}
\setcounter{figure}{0} \renewcommand{\thefigure}{S\arabic{figure}}
\setcounter{table}{0} \renewcommand{\thetable}{S\arabic{table}}
\appendix

\onecolumn
\section*{\Large{Supplementary Material: Are Bayesian neural networks intrinsically good at out-of-distribution detection?}}
\textbf{Christian Henning*, Francesco D'Angelo* and Benjamin F. Grewe}
\section{Additional experiments and results}
In this section, we report additional experiments and results. 
\begin{figure}[h]
     \centering
     \begin{subfigure}[b]{0.24\textwidth}
         \centering
         \includegraphics[width=\textwidth]{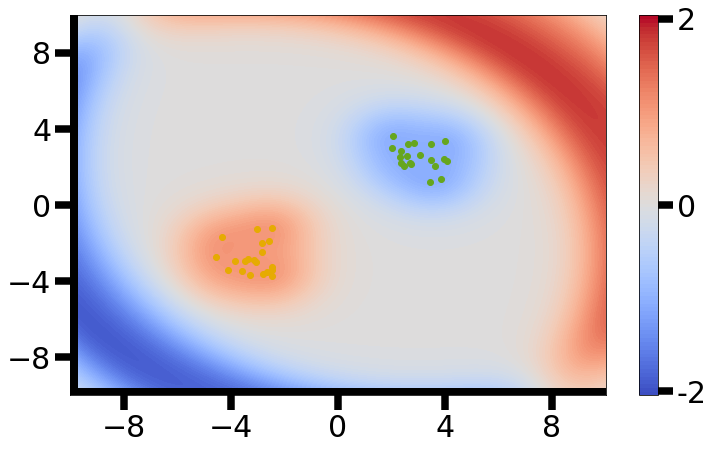}
         \caption{$\bar{\vf}_*$ -- periodic kernel}
         \label{sm:fig:gp:expsinesq:mean}
     \end{subfigure}
     \hfill
     \begin{subfigure}[b]{0.24\textwidth}
         \centering
         \includegraphics[width=\textwidth]{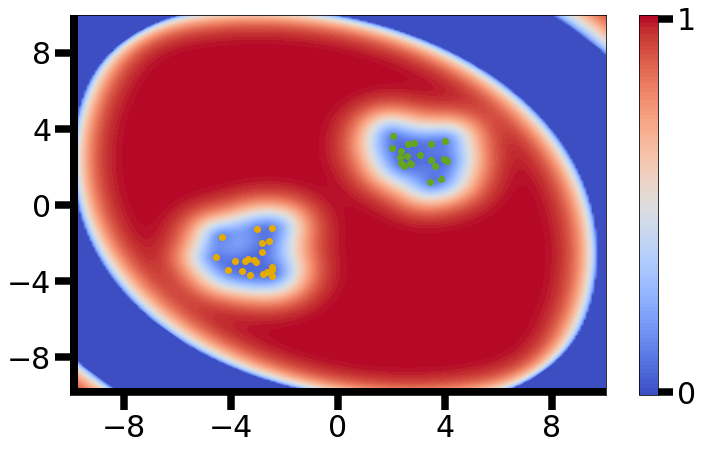}
         \caption{$\sigma(\vf_*)$ -- periodic kernel}
         \label{sm:fig:gp:expsinesq:std}
     \end{subfigure}
     \hfill
     \begin{subfigure}[b]{0.24\textwidth}
         \centering
         \includegraphics[width=\textwidth]{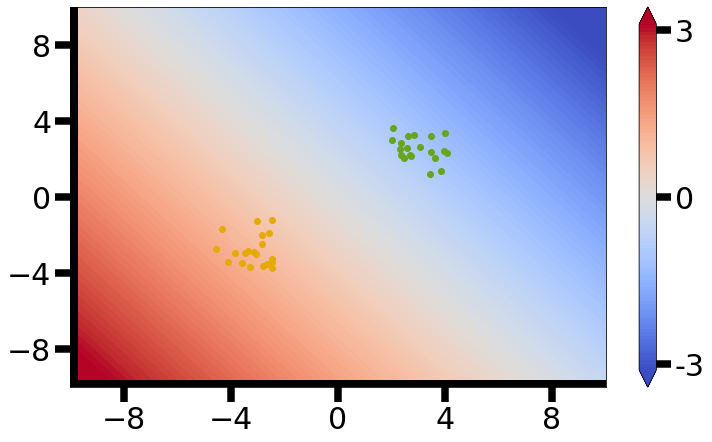}
         \caption{$\bar{\vf}_*$ -- dot-product kernel}
         \label{sm:fig:gp:dot:mean}
     \end{subfigure}
     \begin{subfigure}[b]{0.24\textwidth}
         \centering
         \includegraphics[width=\textwidth]{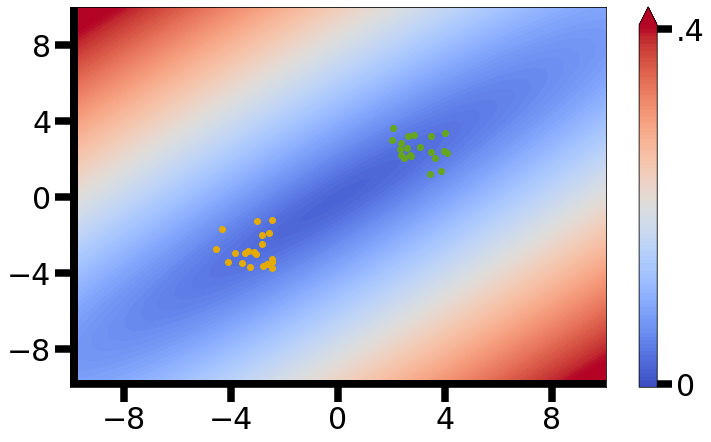}
         \caption{$\sigma(\vf_*)$ -- dot-product kernel}
         \label{sm:fig:gp:dot:std}
     \end{subfigure}
        \caption{Mean $\bar{\vf}_*$ and standard deviation $\sigma(\vf_*)$ of the posterior $p(\vf_* \mid X_*,X, \vy)$ visualized for GPs with traditional kernels. In contrast to a GP with RBF kernel (cf. Fig. \ref{fig:main:ood:plots}), the kernels used here induce epistemic uncertainty $\sigma(\vf_*)$ that is undesirable for the purpose of OOD detection. Therewith, this plot highlights the importance of the function space prior for the task of detecting anomalies using predictive uncertainty.}
        \label{sm:fig:traditional:kernels}
\end{figure}

\begin{figure}[h]
     \centering
     \begin{subfigure}[b]{0.24\textwidth}
         \centering
         \includegraphics[width=\textwidth]{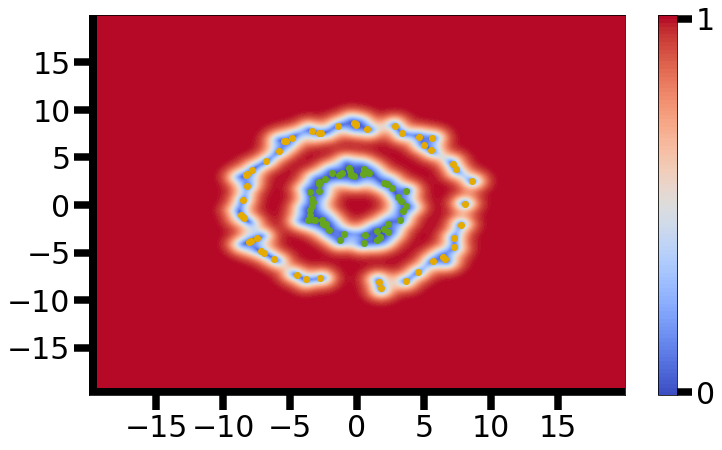}
         \caption{$\sigma(\vf_*)$ -- RBF kernel}
         \label{fig:gp:rbf:std:donuts}
     \end{subfigure}
     \hfill
     \begin{subfigure}[b]{0.24\textwidth}
         \centering
         \includegraphics[width=\textwidth]{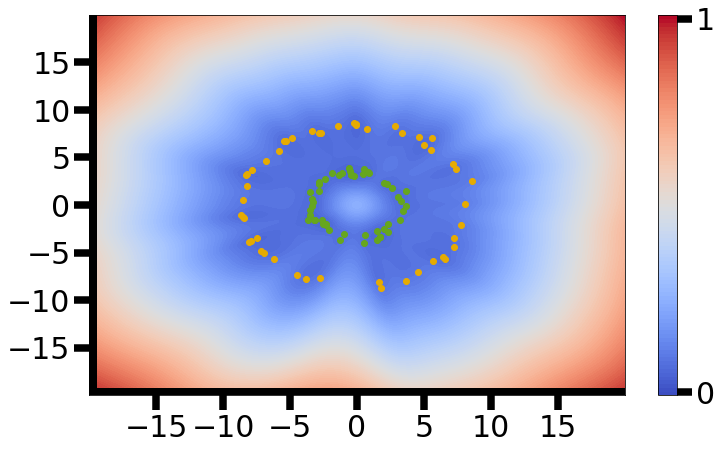}
         \caption{$\sigma(\vf_*)$ -- 2-layer ReLU ($\infty$)}
         \label{fig:gp:ana:relu:nngp:2l:donuts}
     \end{subfigure}
     \hfill
     \begin{subfigure}[b]{0.247\textwidth}
         \centering
         \includegraphics[width=\textwidth]{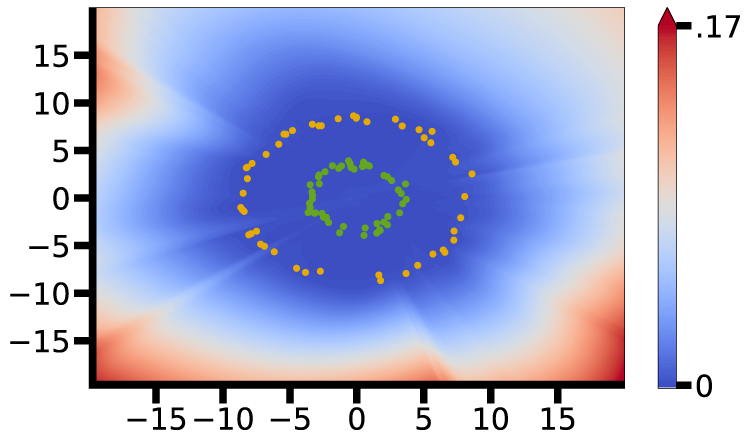}
         \caption{$\sigma(\vf_*)$ -- 2-layer ReLU (5)}
         \label{fig:hmc:std:relu:2l:width:10:donuts}
     \end{subfigure}
     \begin{subfigure}[b]{0.247\textwidth}
         \centering
         \includegraphics[width=\textwidth]{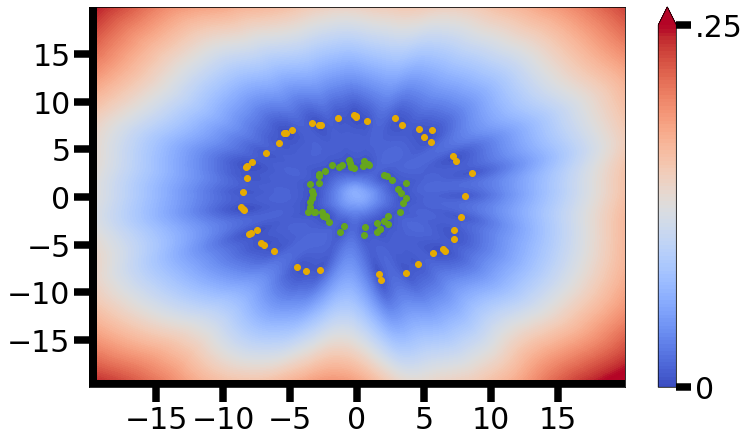}
         \caption{$\sigma(\vf_*)$ -- 2-layer ReLU (100)}
         \label{fig:hmc:std:relu:2l:width:100:donuts}
     \end{subfigure}
    \caption{\textbf{Standard deviation of the predictive posterior.} We perform Bayesian inference on a datasets composed by two concentric rings comparing a GP with RBF kernel and ReLU networks in the finite and infinite regimes (in analogy to Fig. \ref{fig:main:ood:plots}). Only the function space prior induced by an RBF kernel causes epistemic uncertainties that allow outlier detection in the center or in between the two circles.}
    \label{fig:main:ood:plots:donuts}
\end{figure}
\vspace{5mm}
\begin{figure}[H]
     \centering
     \includegraphics[width=0.60\textwidth]{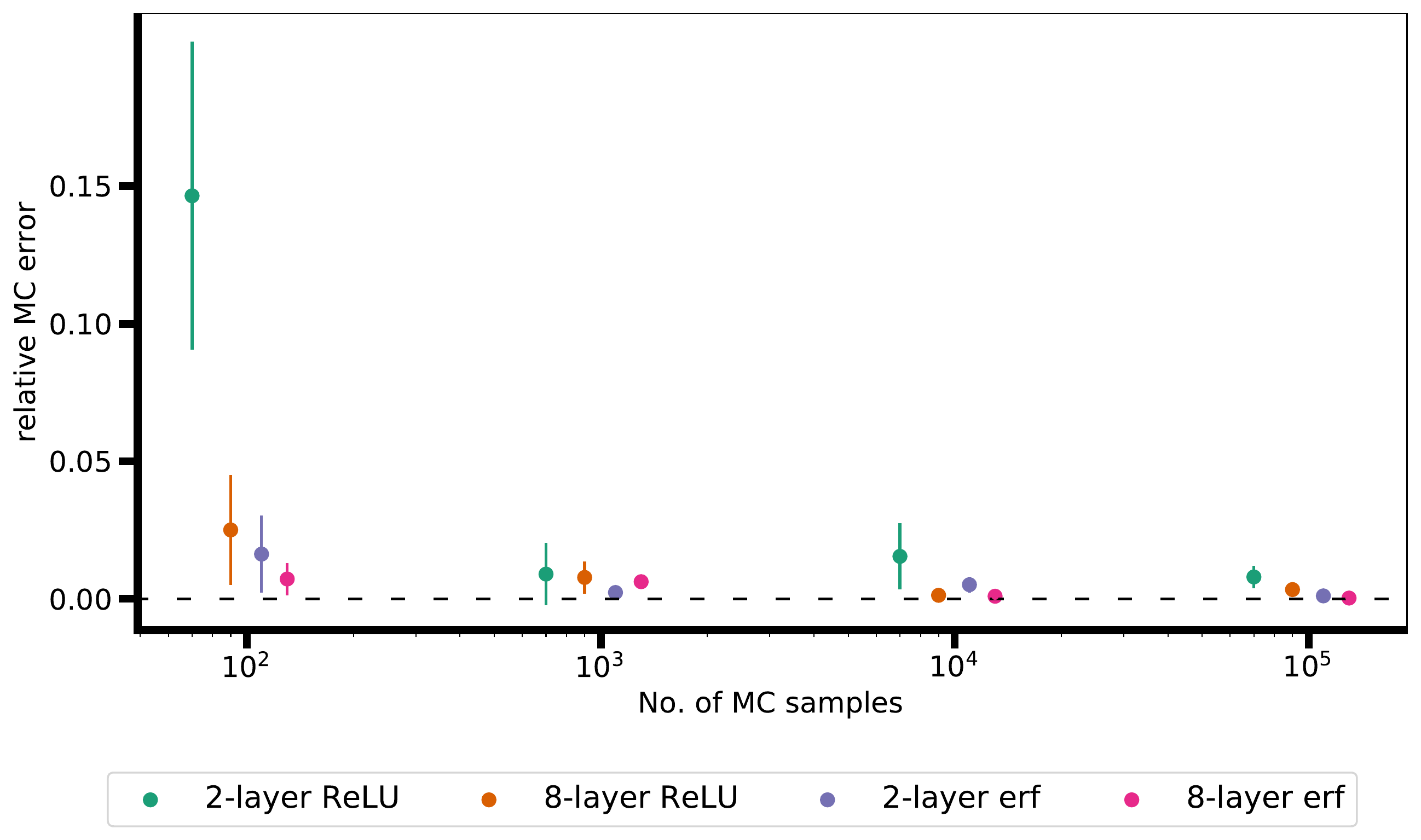}
    \caption{Eq.~\ref{eqn:MC_NNGP} requires estimation whenever no analytic kernel expression is available (for instance, when using a hyberbolic tangent nonlinearity as in Fig.~\ref{sm:fig:nngp:posteriors}). Here, we visualize the error caused by this approximation for ReLU and error function (erf) networks when computing kernel values $k(x_*,x_*)$. Eq.~\ref{eqn:MC_NNGP} requires a recursive estimation of expected values, where we estimate each of them using $N$ samples ($N \in [10^2, 10^3, 10^4, 10^5]$). Note, that even small errors can cause eigenvalues of the kernel matrix to become negative. However, with our chosen likelihood variance of $\sigma = 0.02$ we experience no numerical instabilities during inference, and obtain consistent results using either analytic or estimated ($N=10^5$) kernel matrices.}
    \label{sm:fig:mc:error}
\end{figure}

\begin{figure}[h]
     \centering
     \begin{subfigure}[b]{0.24\textwidth}
         \centering
         \includegraphics[width=\textwidth]{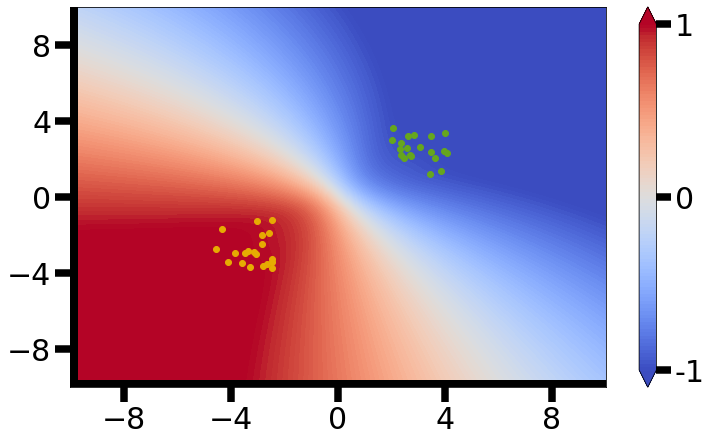}
         \caption{$\bar{\vf}_*$ -- ReLU 1 layer}
         \label{sm:fig:nngp:post:mean:ana:relu:1l}
     \end{subfigure}
     \hfill
     \begin{subfigure}[b]{0.24\textwidth}
         \centering
         \includegraphics[width=\textwidth]{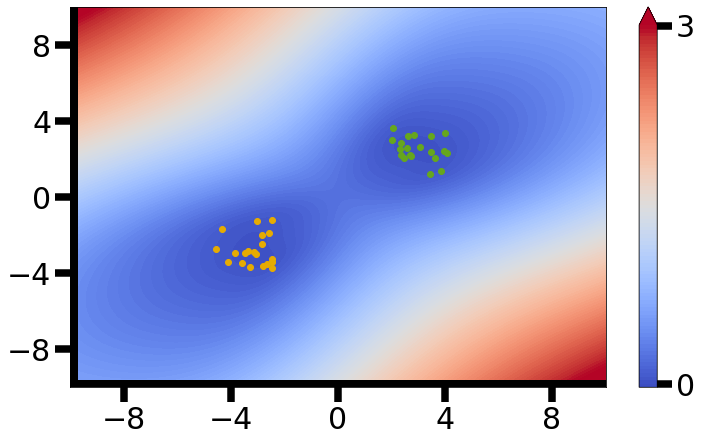}
         \caption{$\sigma(\vf_*)$ -- ReLU 1 layer}
         \label{sm:fig:nngp:post:std:ana:relu:1l}
     \end{subfigure}
     \hfill
     \begin{subfigure}[b]{0.24\textwidth}
         \centering
         \includegraphics[width=\textwidth]{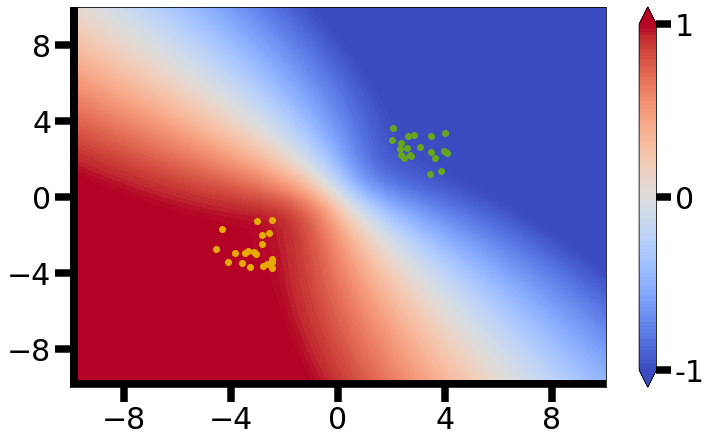}
         \caption{$\bar{\vf}_*$ -- ReLU 2 layer}
         \label{sm:fig:nngp:post:mean:ana:relu:2l}
     \end{subfigure}
     \begin{subfigure}[b]{0.24\textwidth}
         \centering
         \includegraphics[width=\textwidth]{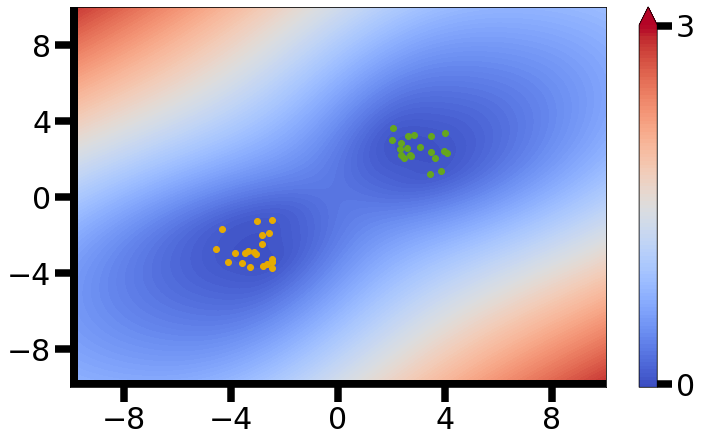}
         \caption{$\sigma(\vf_*)$ -- ReLU 2 layer}
         \label{sm:fig:nngp:post:std:ana:relu:2l}
     \end{subfigure}
     
     \begin{subfigure}[b]{0.24\textwidth}
         \centering
         \includegraphics[width=\textwidth]{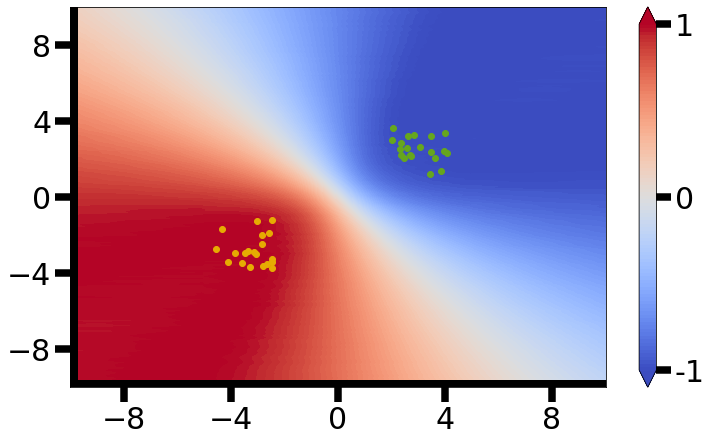}
         \caption{$\bar{\vf}_*$ -- Tanh 1 layer}
         \label{sm:fig:nngp:post:mean:tanh:1l}
     \end{subfigure}
     \hfill
     \begin{subfigure}[b]{0.24\textwidth}
         \centering
         \includegraphics[width=\textwidth]{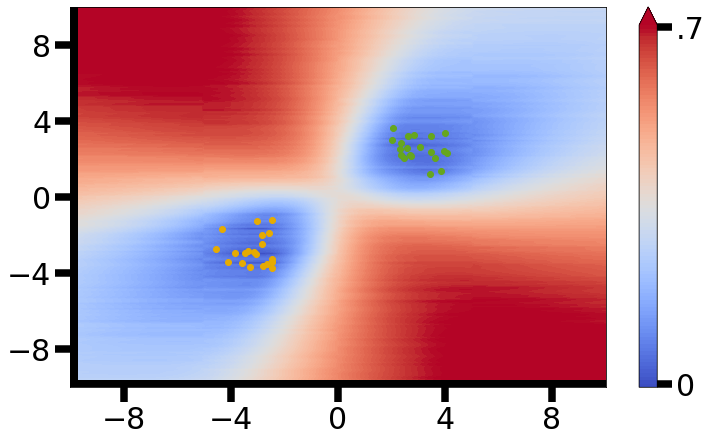}
         \caption{$\sigma(\vf_*)$ -- Tanh 1 layer}
         \label{sm:fig:nngp:post:std:tanh:1l}
     \end{subfigure}
     \hfill
     \begin{subfigure}[b]{0.24\textwidth}
         \centering
         \includegraphics[width=\textwidth]{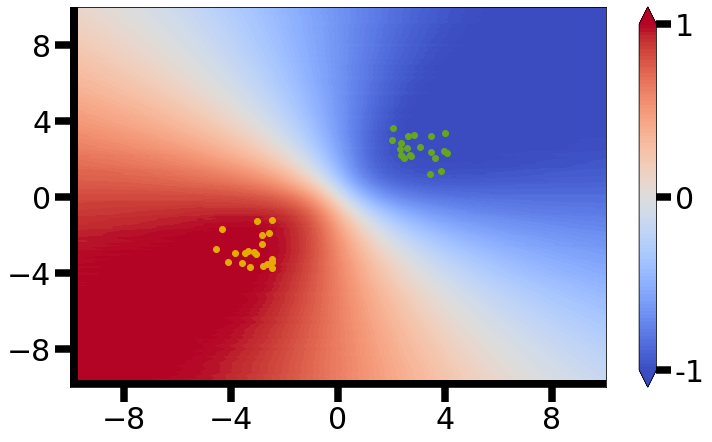}
         \caption{$\bar{\vf}_*$ -- Tanh 2 layer}
         \label{sm:fig:nngp:post:mean:tanh:2l}
     \end{subfigure}
     \begin{subfigure}[b]{0.24\textwidth}
         \centering
         \includegraphics[width=\textwidth]{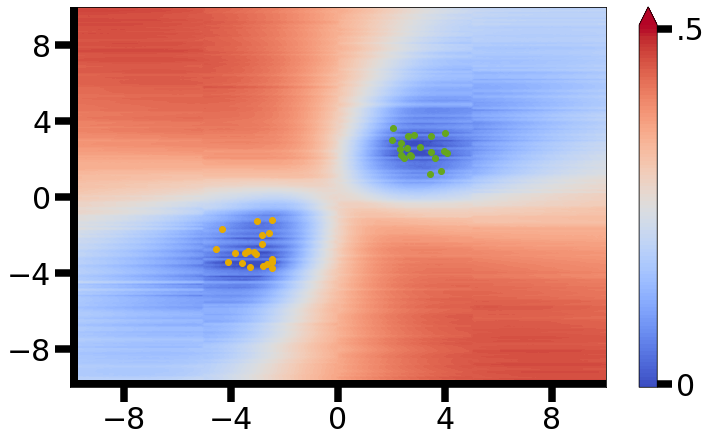}
         \caption{$\sigma(\vf_*)$ -- Tanh 2 layer}
         \label{sm:fig:nngp:post:std:tanh:2l}
     \end{subfigure}
     
          \begin{subfigure}[b]{0.24\textwidth}
         \centering
         \includegraphics[width=\textwidth]{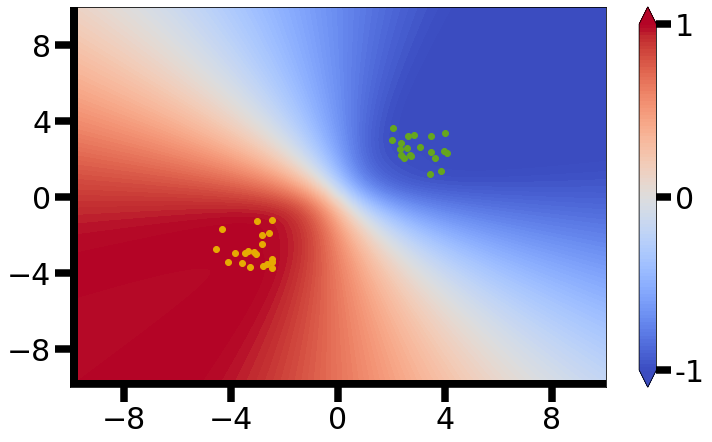}
         \caption{$\bar{\vf}_*$ -- Erf 1 layer}
         \label{sm:fig:nngp:post:mean:erf:1l}
     \end{subfigure}
     \hfill
     \begin{subfigure}[b]{0.24\textwidth}
         \centering
         \includegraphics[width=\textwidth]{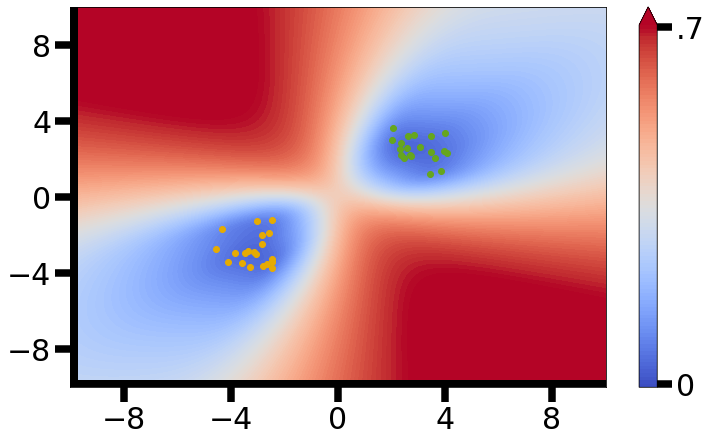}
         \caption{$\sigma(\vf_*)$ -- Erf 1 layer}
         \label{sm:fig:nngp:post:std:erf:1l}
     \end{subfigure}
     \hfill
     \begin{subfigure}[b]{0.24\textwidth}
         \centering
         \includegraphics[width=\textwidth]{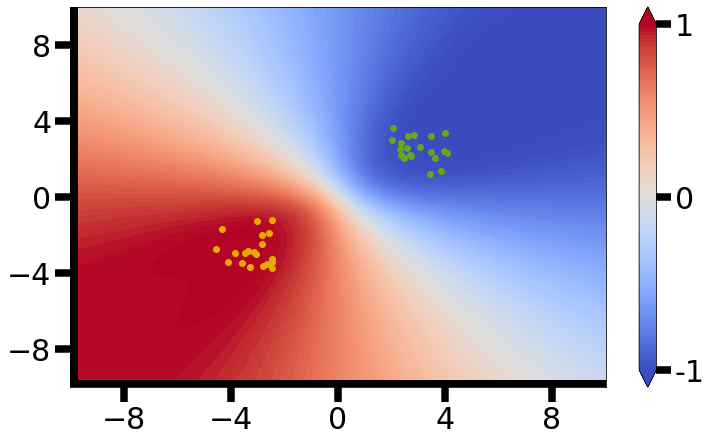}
         \caption{$\bar{\vf}_*$ -- Erf 2 layer}
         \label{sm:fig:nngp:post:mean:erf:2l}
     \end{subfigure}
     \begin{subfigure}[b]{0.24\textwidth}
         \centering
         \includegraphics[width=\textwidth]{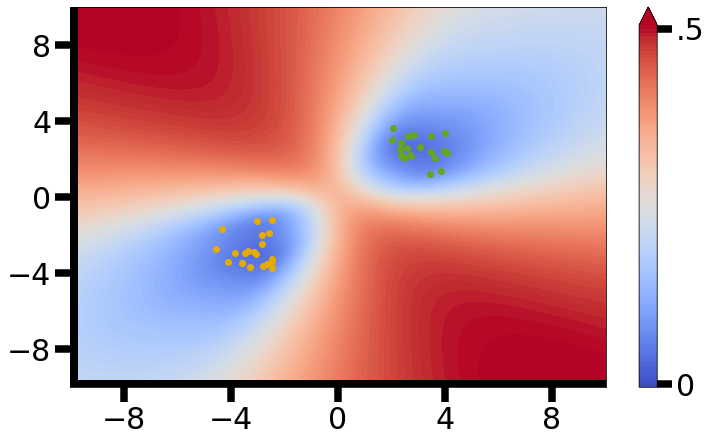}
         \caption{$\sigma(\vf_*)$ -- Erf 2 layer}
         \label{sm:fig:nngp:post:std:erf:2l}
     \end{subfigure}
    \caption{Mean $\bar{\vf}_*$ and standard deviation $\sigma(\vf_*)$ of the posterior $p(\vf_* \mid X_*,X, \vy)$ visualized for GP regression using NNGP kernels for various architectural choices, such as number of layers or non-linearity. Note, that Tanh and Erf nonlinearities are quite similar in shape, which is reflected in the similar predictive posterior that is induced by these networks. We use the analytically known kernel expression for the Erf kernel \cite{williams1997rbf}, and use MC sampling for the Tanh network.}
    \label{sm:fig:nngp:posteriors}
\end{figure}

\begin{figure}[H]
     \centering
     \includegraphics[width=0.80\textwidth]{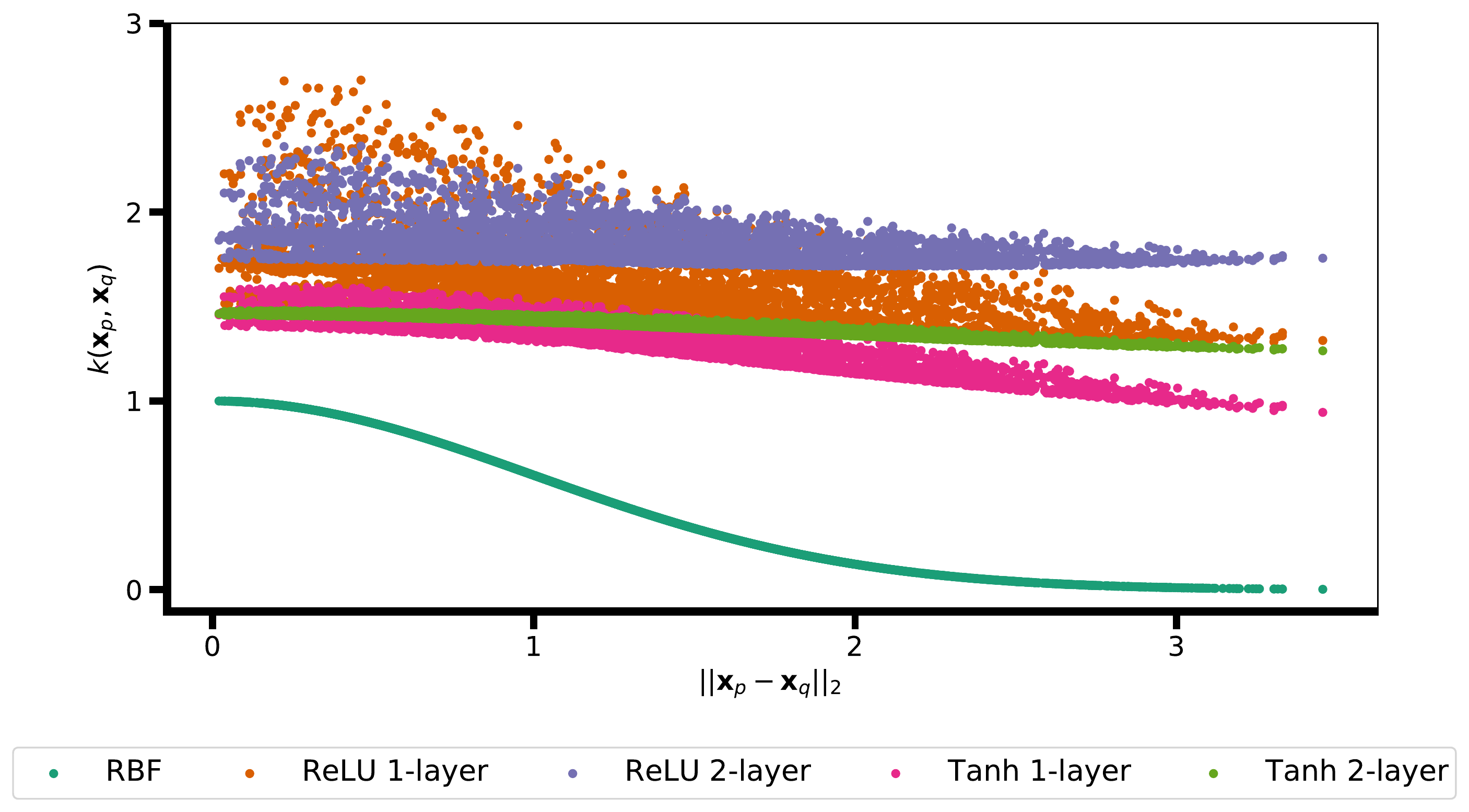}
    \caption{Kernel values plotted as a function of the Euclidean distance of points (using pairs of training points from the two Gaussian mixtures dataset). As outlined in Sec.~\ref{sec:discussion} and Sec.~\ref{sm:sec:rbf:kde}, the interpretation of an RBF kernel as Gaussian kernel that can be used in a KDE of $p(\vx)$ is important to justify implied OOD capabilities. Unfortunately, the kernels induced by common architectures do not seem to be distance-aware and are thus not useful for KDE.}
    \label{sm:fig:nngp:distance:awareness}
\end{figure}

\begin{figure}[H]
    \centering
    \begin{subfigure}[b]{0.24\textwidth}
        \centering
        \includegraphics[width=\textwidth]{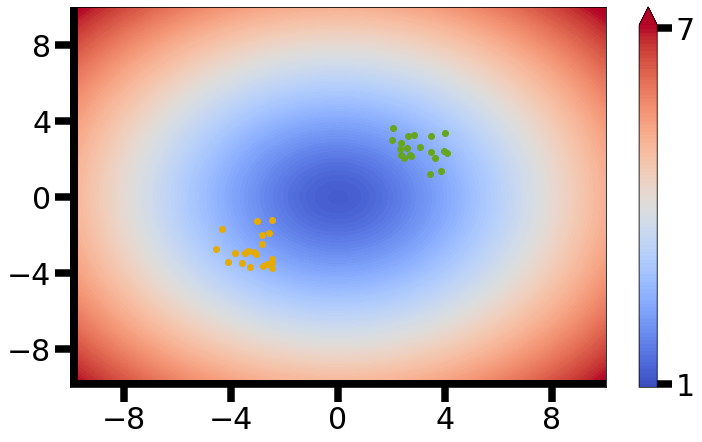}
        \caption{$k(x_*,x_*)$ -- ReLU 1 layer}
        \label{sm:fig:nngp:prior:relu:1l}
    \end{subfigure}
    \hfill
    \begin{subfigure}[b]{0.24\textwidth}
        \centering
        \includegraphics[width=\textwidth]{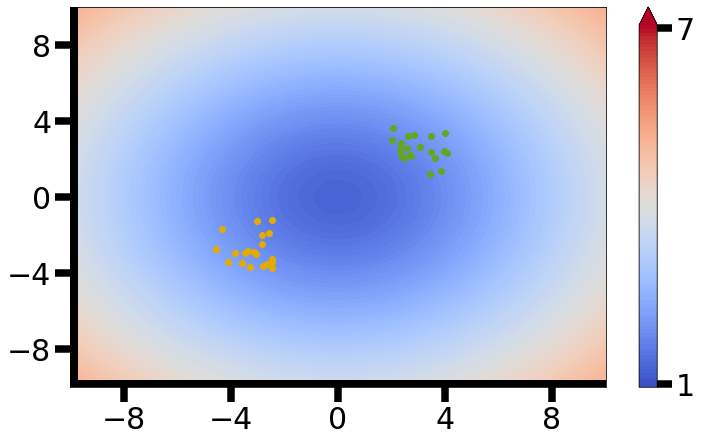}
        \caption{$k(x_*,x_*)$ -- ReLU 2 layer}
        \label{sm:fig:nngp:prior:relu:2l}
    \end{subfigure}
    \hfill
    \begin{subfigure}[b]{0.247\textwidth}
        \centering
        \includegraphics[width=\textwidth]{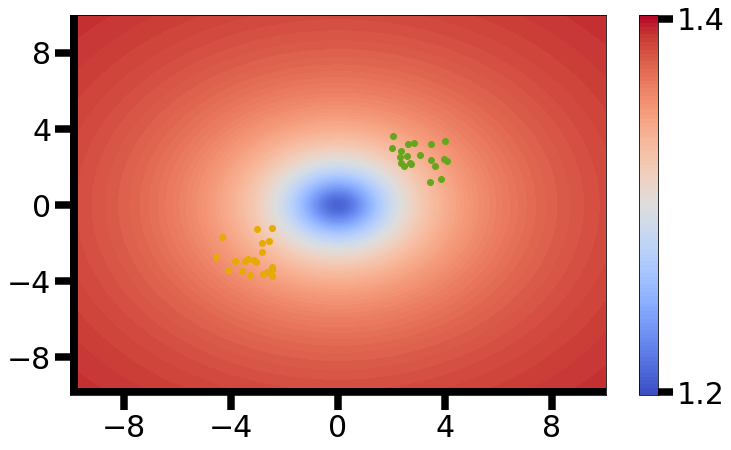}
        \caption{$k(x_*,x_*)$ -- Erf 1 layer}
        \label{sm:fig:nngp:prior:erf:1l}
    \end{subfigure}
    \begin{subfigure}[b]{0.247\textwidth}
        \centering
        \includegraphics[width=\textwidth]{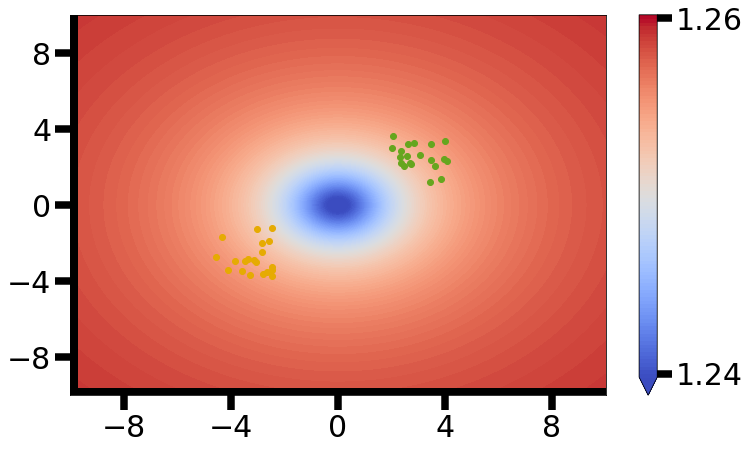}
        \caption{$k(x_*,x_*)$ -- Erf 2 layer}
        \label{sm:fig:nngp:prior:erf:2l}
    \end{subfigure}
    \caption{NNGP kernel values for various architectural choices. Note, that the NNGP kernel value $k(x_*,x_*)$ represents the prior variance of function values under the induced GP prior at the location $x_*$ (cf. Eq.~\ref{eq:gp:prior}). As emphasized in the main text, $k(x_*,x_*)$ is constant for an RBF kernel, which has important implications for OOD detection, such as that a priori (before seeing any data) all points are treated equally. This is not the case for the Relu Kernel, which has an angular dependence and depends on the input norm (cf. Eq.~\ref{eqn:k:recursion:start} and Eq.~\ref{eqn:k_relu}). The kernel induced by networks using a error function (Erf) as nonlinearity seems to be more desirable in this respect (note the scale of the colorbars).}
    \label{sm:fig:prior:kernels}
\end{figure}

\begin{figure}[H]
     \centering
     \begin{subfigure}[b]{0.45\textwidth}
         \centering
         \includegraphics[width=\textwidth]{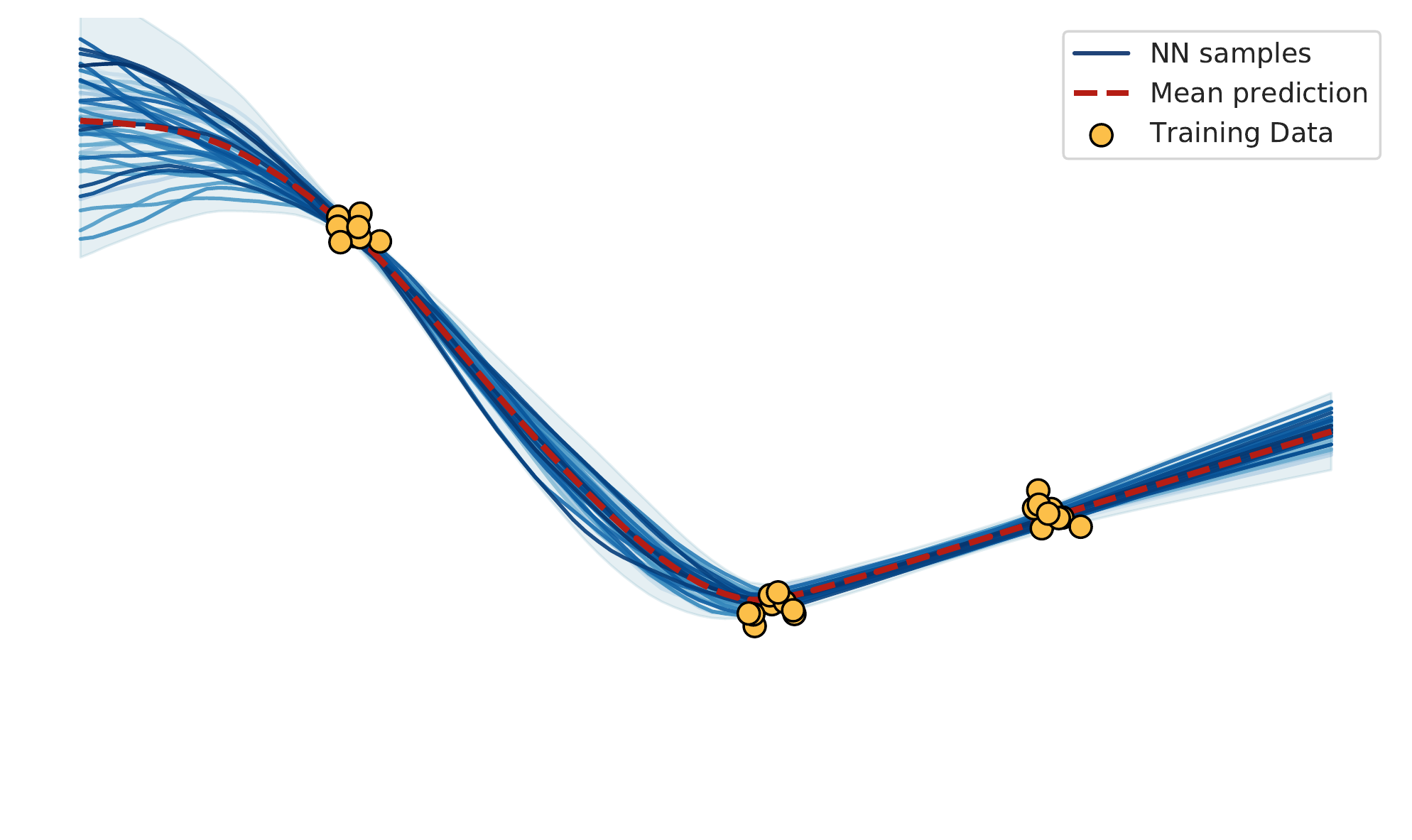}
         \caption{Width-aware prior}
         \label{sm:fig:scaled:prior}
     \end{subfigure}
     \begin{subfigure}[b]{0.45\textwidth}
         \centering
         \includegraphics[width=\textwidth]{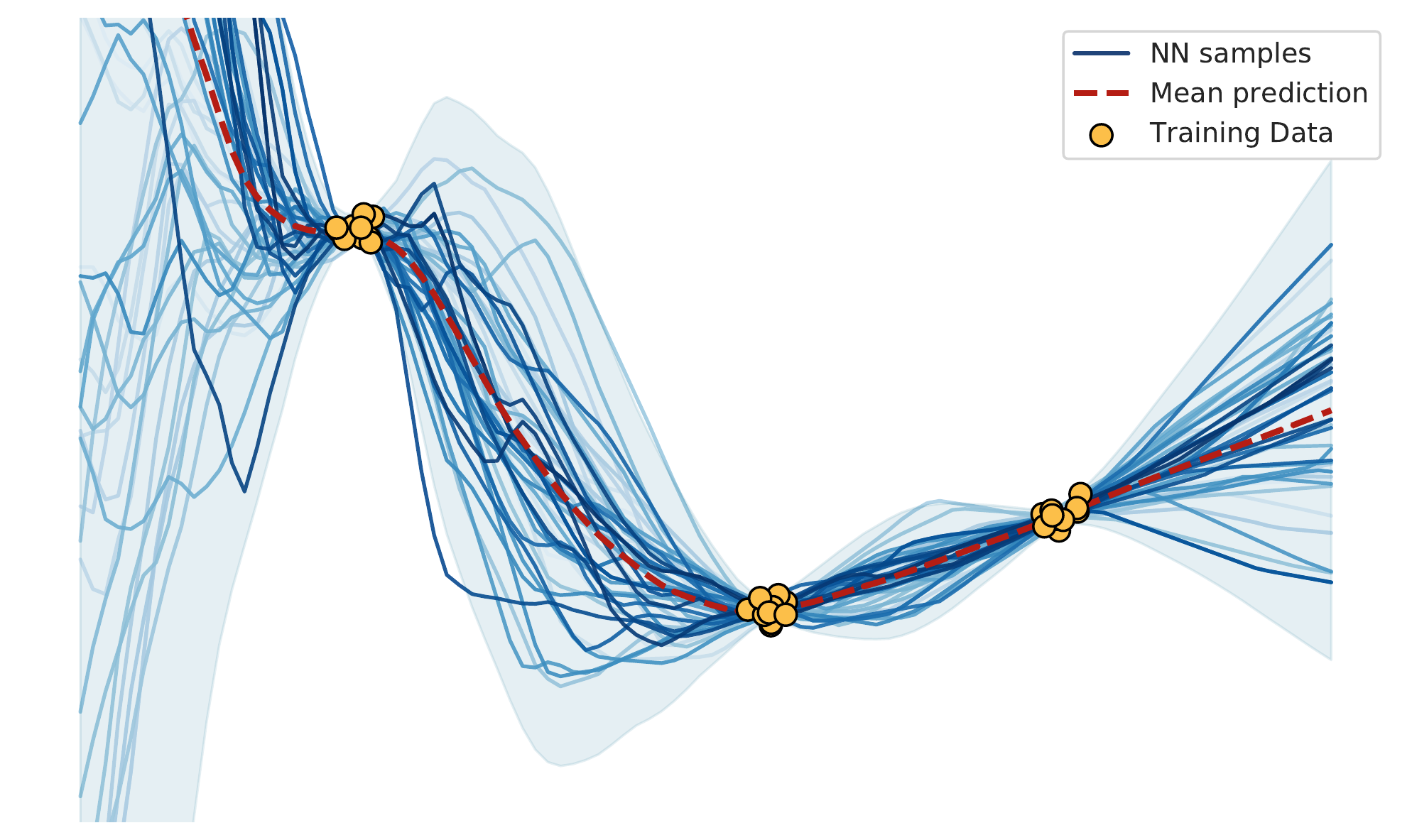}
         \caption{Standard prior}
         \label{sm:fig:stadard:prior}
     \end{subfigure}
         \caption{This figure illustrates the importance of the choice of weight prior $p(\vw)$. Here, we perform 1d regression using HMC with either a width-aware prior $\mathcal{N}(0,\frac{\sigma_w^2}{H_l})$ or a standard prior $\mathcal{N}(0,\sigma_w^2)$.}
        \label{sm:fig:prior:importance}
\end{figure}

\section{What is an out-of-distribution input?}
\label{sm:sec:ood}

The corresponding Wikipedia article uses the following formulation: "Determining whether or not an observation is an outlier is ultimately a subjective exercise" \cite{zimek2018outlier:detection}.\footnote{\url{https://en.wikipedia.org/wiki/Outlier}, accessed: 2021-06-01} \citet{pimentel2014review:novelty} reviews methods for outlier detection, putting them coarsely into 5 categories: (1) probabilistic, (2) distance-based, (3) reconstruction-based, (4) domain-based, and (5) information-theoretic. Our focus lies on a probabilistic characterization of an OOD point, where a statistical criterion allows to decide whether a given input is significantly different from the observed training population.\footnote{Note, that we are not considering the equally important problem of outlier detection within the training set \cite{zimek2018outlier:detection}.} In this section, we discuss pitfalls regarding obvious choices for such a criterion.

We would like to start by emphasizing why a mathematical characterization of "out of distribution" is desirable. As indicated in Sec.~\ref{sec:intro}, applying Bayesian statistics to neural networks should ultimately serve the goal of providing safety guarantees for critical applications. Any such guarantee will rely on assumptions (e.g., the assumption that the neural network architecture is sufficient to encompass the ground-truth function). However, it is important to be aware of any such assumptions, such that their validity can be tested. The scope of this work is meant to encourage research towards design principles that facilitate BNNs that are more reliable at OOD detection. We argue that the effectiveness of such BNNs can only be thoroughly examined if the concept of OOD is clearly defined. 

There are many definitions that could be considered for a point to be OOD as we will outline below. Any of these definitions may change the notion of OOD for individual inputs and will therefore affect how a BNN should be designed such that predictive uncertainty adheres to the underlying OOD definition. Considering a generative process $p(\vx)$, the first question arising is regarding the regions outside the support of $p(\vx)$ or even the space outside the manifold where samples $\vx$ are defined on. For instance, assume the data to be images embedded on a lower-dimensional manifold. Are points outside this manifold clearly OOD, given that minor noise corruptions are likely to leave the manifold? Even disregarding these topological issues solely focusing on the density $p(\vx)$, makes the distinction between in- and out-of-distribution challenging. For instance, a threshold-criterion on the density might cause samples in a zero probability region to be considered in-distribution \cite{nalisnick:2018ood:generative}. To overcome such challenges, one may resort to concepts from information theory, such as the notion of a typical set \cite{mackay:information:theory, nalisnick2019typical}. Unfortunately, an OOD criterion based on this notion would require looking at sets rather than individual points. We hope that this short outline highlights the challenges regarding the definition of OOD, but also clarifies that a proper definition is relevant when assessing OOD capabilities on high-dimensional data, where a visual assessment as in Fig.~\ref{fig:main:ood:plots} is not possible.


\section{Note on the relation between GP regression and kernel density estimation}
\label{sm:sec:rbf:kde}

GP regression with an RBF kernel allows a direct understanding of the OOD capabilities that a Bayesian posterior may possess, as the a priori uniform epistemic uncertainty is reduced in direct correspondence to density of $p(\vx)$.

Below, we rewrite the variance of an input $x_*$ as defined in Eq.~\ref{eqn:gp_regr}:

\begin{equation}
        \sigma^2(\vf_*) = k(x_*,x_*) - \sum_{i=1}^n \beta_i(x_*) k(x_*,x_i)
\end{equation}

with  $\beta_i(x_*) = \sum_{j=1}^n \left( K(X,X) + \sigma^2 \mathbb{I} \right)^{-1}_{ij} k(x_*,x_j)$.

Note, that $k(x_*,x_*)$ is a positive constant for an RBF kernel, and that $\beta_i(x_*) \geq 0$. Furthermore, as the RBF kernel $k(x_*,x_j)$ behaves exponentially inverse to the distance between $x_*$ and $x_j$, $\beta_i(x_*)$ is approximately zero for all $x_*$ that are far from all training points. In addition, a training point $x_i$ can only decrease the prior variance if it is close to $x_*$. This analogy closely resembles the philosophy of KDE, which becomes an exact generative model in the limit of infinite data and bandwidth $l \rightarrow 0$.

\section{The kernel induced by RBF networks}
\label{sm:sec:rbf:nets}

\begin{figure*}
     \centering
     \begin{subfigure}[b]{0.24\textwidth}
         \centering
         \includegraphics[width=\textwidth]{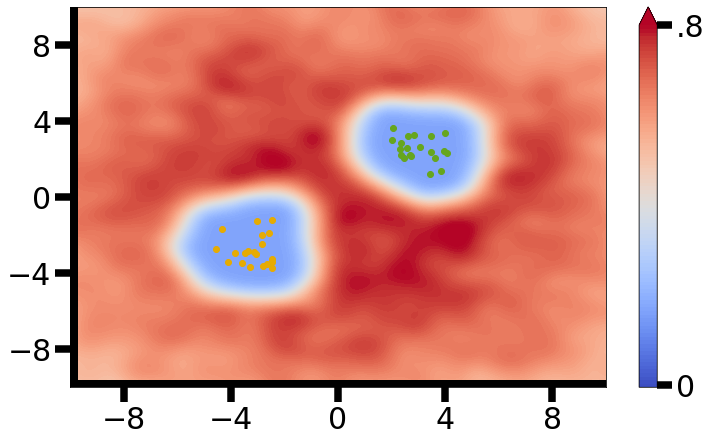}
         \caption{$\sigma(\vf_*)$ -- RBF net (500)}
         \label{sm:hmc:RBF_net:g_mix}
     \end{subfigure}
     \hfill
     \begin{subfigure}[b]{0.24\textwidth}
         \centering
         \includegraphics[width=\textwidth]{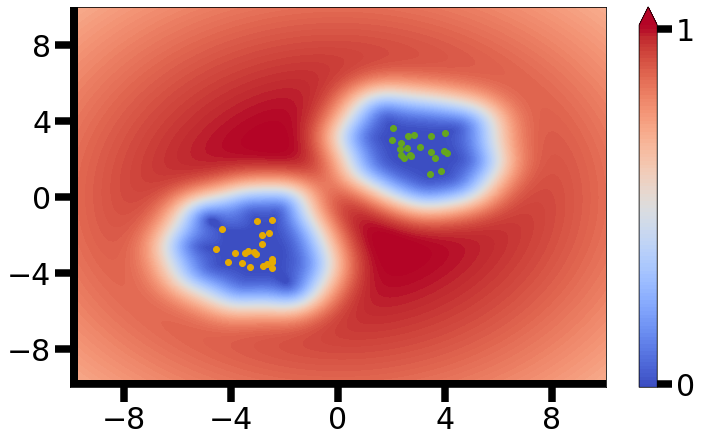}
         \caption{$\sigma(\vf_*)$ -- RBF net ($\infty$)}
         \label{sm:gp:RBF_net:g_mix}
     \end{subfigure}
     \hfill
     \begin{subfigure}[b]{0.247\textwidth}
         \centering
         \includegraphics[width=\textwidth]{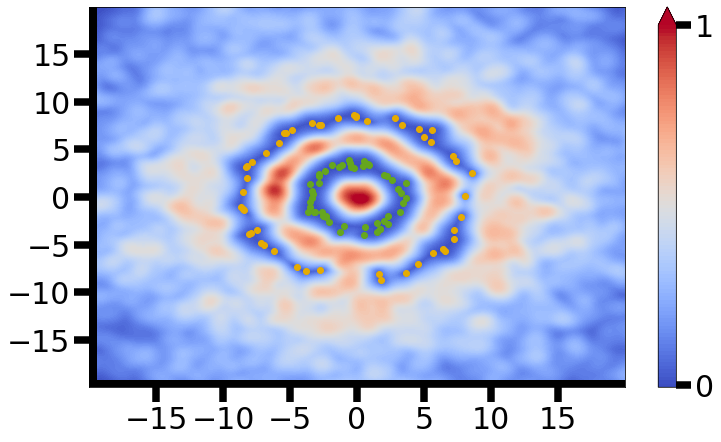}
         \caption{$\sigma(\vf_*)$ -- RBF net (500)}
         \label{sm:hmc:RBF_net:donuts}
     \end{subfigure}
     \begin{subfigure}[b]{0.247\textwidth}
         \centering
         \includegraphics[width=\textwidth]{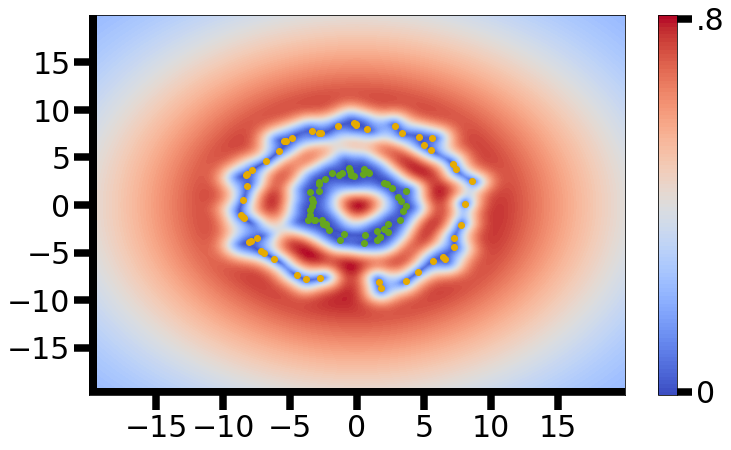}
         \caption{$\sigma(\vf_*)$ -- RBF net ($\infty$)}
         \label{sm:gp:RBF_net:donuts}
     \end{subfigure}
        \caption{Standard deviation $\sigma(\vf_*)$ of the posterior $p(\vf_* \mid X_*,X, \vy)$ visualized for a one hidden-layer RBF network with 500 hidden units (a,c) and the correspondent infinite GP (b,d). $\sigma(\vf_*)$ has been estimated using HMC for the finite cases.}
        \label{sm:fig:rbf:networks}
\end{figure*}
Commonly, we consider neural networks where computational units (neurons) apply an affine transformation to their inputs followed by a non-linear function. Another class of neural networks are formed by radial basis function (RBF) networks, which further enhance the connection with Gaussian processes. Indeed the use of the exponential activation function, while being an uncommon choice in the deep learning community, can offer intriguing properties for OOD detection that closely resemble the ones of a GP with RBF kernel. In particular, units in a finite RBF network perform a linear combination of radial basis functions: 
\begin{equation*}
    f(\vx) = \sum_{j=1}^H w_j \exp \left( - \frac{1}{2\sigma^2_g}\lVert \vx - \boldsymbol{\mu}_j \rVert^2 \right) + b
\end{equation*}
where the centers $\boldsymbol{\mu}_j$, the bias $b$ and the coefficients $w_j$ are learned. Analogous to the NNGP case in Sec. \ref{sec:background} such networks converge to a GP with $f(\vx) \sim \mathcal{GP}(0,C)$ in the infinite-width case when assuming a Gaussian prior over the centers $\boldsymbol{\mu}_j \sim \mathcal{N}(0,\sigma_\mu^2 \mathbb{I})$ and $w_j \sim \mathcal{N}(0,\sigma_w^2)$. An analytical expression for the correspondent kernel \cite{williams1997rbf} in the case of a single hidden layer is given by: 

\begin{align*}
    C(\vx,\vx') &= \overbrace{\mathbb{E} \left[f^l(\vx) f^l(\vx') \right]}^{k^l(\vx,\vx')} - \overbrace{\mathbb{E} \left[f^l(\vx)\right]  \mathbb{E} \left[f^l(\vx)\right]}^{=0} \\
    &= \sigma_b^2 + \sigma^2_w \mathbb{E} \left[\exp\left( - \frac{1}{2\sigma^2_g}\lVert \vx - \boldsymbol{\mu}_j \rVert^2 \right)  \exp\left( - \frac{1}{2\sigma^2_g}\lVert \vx' - \boldsymbol{\mu}_j \rVert^2 \right)  \right]
\end{align*}
Where the last expectation has a closed form solution so that: 
\begin{align*}
    C(\vx,\vx') =\sigma_b^2 + \sigma_w^2 \bigg( \frac{\sigma_e}{\sigma_\mu}\bigg)^d \exp \bigg(- \frac{\lVert \vx \rVert^2}{2 \sigma^2_m} \bigg) \exp \bigg( -\frac{\lVert \vx - \vx' \rVert^2 }{{2 \sigma^2_s}} \bigg) & \exp \bigg(- \frac{\lVert \vx'\rVert^2}{2 \sigma^2_m} \bigg)
\end{align*}
where $d$ is the dimension of the data $\vx$, $1/\sigma_e^2 = 2/\sigma_g^2 + 1/\sigma^2_\mu$, $\sigma_s^2 = 2 \sigma^2_g + \sigma^4_g/\sigma_\mu^2$ and $\sigma_m^2 = 2\sigma_\mu^2 + \sigma_g^2$.
It is important to notice that there is an equivalence with the RBF kernel up to a constant when the inputs have constant norm. In addition, the kernel becomes proportional to the RBF kernel in the limit $\sigma_\mu^2 \rightarrow \infty$ \cite{williams1997rbf}.

We tested the OOD performance of a finite-width RBF neural network in Fig. \ref{sm:hmc:RBF_net:g_mix} and \ref{sm:hmc:RBF_net:donuts} using HMC and with $\sigma_\mu^2 = 100$. For moderate norms in input space, the OOD behavior closely assimilates the behavior obtained when using an RBF kernel as in Fig. \ref{fig:gp:rbf:std}. Crucially, the network is able to detect OOD points in the center and in-between region (in contrast to common neural network architectures). The OOD behavior for inputs with a large norm can be improved by further increasing $\sigma_\mu^2$. We also report the correspondent infinite-width limit Fig. \ref{sm:gp:RBF_net:donuts} and \ref{sm:gp:RBF_net:g_mix}. 

\section{Sampling in-distribution data via epistemic uncertainty}

Relating the epistemic uncertainty induced by a BNN to $p(\vx)$ opens up interesting new possibilities. As we have shown in Sec. \ref{sm:sec:rbf:kde}, the epistemic uncertainty of a GP with RBF kernel can be related to a KDE approximation of $ p(\vx)$. Therefore, one may consider the epistemic uncertainty as an energy function to create a generative model from which to sample from the input space.

Moreover, this approach may be used to empirically validate the OOD capabilities of a BNN. Indeed, OOD performance is commonly validated by selecting a specific set of known OOD datasets. However, as the OOD space comprises everything except the in-distribution data, it is infeasible to gain good coverage on high-dimensional data with such testing approach. We therefore suggest a reverse approach that checks the consistency of certainty regions instead of the uncertain ones by sampling via the epistemic uncertainty. Indeed, if these samples and thus the generative model based on the uncertainty are consistent with the in-distribution data and their distribution then a strong indication for trustworthiness is provided. More specifically a sample-based discrepancy measure \cite{kernalized:stein:discrepancy, kernel:two:sample:test} can be used to compare the generated samples and training data points and quantitatively assess like this to which extent the epistemic uncertainty is related to $p(x)$. 

\section{Experimental details}
In this section, we report details on our implementation for the experiments conducted in this work. In all two-dimensional experiments: the likelihood is fixed to be Gaussian with variance $0.02$.  Unless noted otherwise, the prior in weight space is always a width-aware centered Gaussian with $\sigma_w = 1.0$ and $\sigma_b = 1.0$ except for the experiments involving RBF networks where we select $\sigma_w = 200.0$. The RBF kernel bandwidth was fixed to $1.0$ in all corresponding experiments. 
\paragraph{Gaussian Mixture}We created a two-dimensional mixture of two Gaussians with means $\mu_1 = (-2,-2)$, $\mu_2 = (2,2)$ and covariance $\Sigma = 0.5 \cdot \mathbb{I}$ and sampled 20 training datapoints. 
\paragraph{Two rings}We uniformly sampled 50 training datapoints from two rings centred in $(0,0)$ with inner and outer radii $R_{i,1} = 3, R_{o,1} = 4$, $R_{i,2} = 8, R_{o,2} = 9$, respectively.

\paragraph{HMC} We use 5 parallel chains for 5000 steps with each constituting 50 leapfrog steps with stepsize $0.001$ for width 5 and $0.0001$ for width 100. We considered a burn-in phase of 1000 steps and collected a total of 1000 samples.

\end{document}